\documentclass[pdflatex,sn-basic]{sn-jnl}
\usepackage[utf8]{inputenc} 
\usepackage[rightcaption]{sidecap}
\usepackage{booktabs}
\usepackage{multirow}
\usepackage{pdflscape}
\usepackage{hyperref}
\usepackage{graphicx} 
\usepackage{xcolor} 
\usepackage{soul} 
\usepackage{subcaption} 
\usepackage{algorithm}
\usepackage{array}

\usepackage[T1]{fontenc} 

\usepackage{etoolbox}
\makeatletter
\patchcmd{\ps@headings}
{\hbox to \hsize{\hfill Springer Nature 2021 \LaTeX\ template\hfill}}
{\hbox to \hsize{\hfill \hfill}}
{}
{}
\patchcmd{\ps@titlepage}
{\hbox to \hsize{\hfill Springer Nature 2021 \LaTeX\ template\hfill}}
{\hbox to \hsize{\hfill  \hfill}}
{}
{}
\makeatother

\graphicspath{ {images/}, {bst/} }

\jyear{2022}%

\begin{document}

\title[ABRAHAM]{A biologically-inspired multi-modal evaluation of molecular generative machine learning}

\author[1,2]{\fnm{Elizaveta} \sur{Vinogradova}}
\equalcont{These authors contributed equally to this work.}

\author[3]{\fnm{Abay} \sur{Artykbayev}}
\equalcont{These authors contributed equally to this work.}

\author[3]{\fnm{Alisher} \sur{Amanatay}}
\equalcont{These authors contributed equally to this work.}

\author[3]{\fnm{Mukhamejan} \sur{Karatayev}}
\author[3]{\fnm{Maxim} \sur{Mametkulov}}
\author[3]{\fnm{Albina} \sur{Li}}
\author[3]{\fnm{Anuar} \sur{Suleimenov}}
\author[3]{\fnm{Abylay} \sur{Salimzhanov}}
\author[1,2]{\fnm{Karina} \sur{Pats}}
\author[3]{\fnm{Rustam} \sur{Zhumagambetov}}
\author*[1]{\fnm{Ferdinand} \sur{Moln\'ar},\email{ferdinand.molnar@nu.edu.kz}}
\author*[4]{\fnm{Vsevolod} \sur{Peshkov},\email{vsevolod.peshkov@nu.edu.kz}}
\author*[3]{\fnm{Siamac} \sur{Fazli},\email{siamac.fazli@nu.edu.kz}}

\affil*[1]{\orgdiv{Department of Biology}, \orgname{Nazarbayev University}, \city{Nur-Sultan}, \country{Kazakhstan}}
\affil*[2]{\orgdiv{Computer Technologies Laboratory}, \orgname{ITMO University}, \city{St. Petersburg}, \country{Russia}}
\affil*[3]{\orgdiv{Department of Computer Science}, \orgname{Nazarbayev University}, \city{Nur-Sultan}, \country{Kazakhstan}}
\affil*[4]{\orgdiv{Department of Chemistry}, \orgname{Nazarbayev University}, \city{Nur-Sultan}, \country{Kazakhstan}}

\abstract{
	While generative models have recently become ubiquitous in many scientific areas, less attention has been paid to their evaluation. For molecular generative models, the state-of-the-art examines their output in isolation or in relation to its input. However, their biological and functional properties, such as ligand-target interaction is not being addressed. In this study, a novel biologically-inspired benchmark for the evaluation of molecular generative models is proposed.
	Specifically, three diverse reference datasets are designed and a set of metrics are introduced which are directly relevant to the drug discovery process. In particular we propose a recreation metric, apply drug-target affinity prediction and molecular docking as complementary techniques for the evaluation of generative outputs.

	While all three metrics show consistent results across the tested generative models, a more detailed comparison of drug-target affinity binding and molecular docking scores revealed that unimodal predictiors can lead to erroneous conclusions about target binding on a molecular level and a multi-modal approach is thus preferrable.
	The key advantage of this framework is that it incorporates prior physico-chemical domain knowledge into the benchmarking process by focusing explicitly on ligand-target interactions and thus creating a highly efficient tool not only for evaluating molecular generative outputs in particular, but also for enriching the drug discovery process in general.
}


\maketitle

\section{Introduction}\label{sec:Introduction}

Drug discovery and development is a highly cost-intensive and time consuming process that may take decades and according to some estimates requires up to \$2.6 billion~\cite{drug_discovery_Hughes} per marketed drug. One of the main difficulties of drug discovery is related to the size of the so-called chemical space, i.e. the space of all theoretically possible molecules adhering to the physico-chemical rules. Even after limiting this space to all molecules obeying Lipinski’s ``rule of 5''~\cite{LIPINSKI19973}, its size is still in the range of 10\textsuperscript{23}-10\textsuperscript{60}~\cite{Reymond2015Chemical}, which is so vast that an exhaustive search for new molecules is practically impossible. Thus, a wide range of generative models for drug discovery have recently been proposed and attracted the attention of the scientific community~\cite{gebauerNIPS,Segler, Kadurin, druGAN, Prykhodko2019, Maziarka2020, transmol_D1RA03086H, ligdream_skalic2019shape, cdn_10.1145/3219819.3219882, reinvent, reinvent3, Moret2020, TransVAE, zhuma2021}. Such models can explore and navigate the chemical space efficiently and thereby facilitate the discovery and generation of new drugs.

To standardize and evaluate the performance of these generative models, a number of benchmarking metrics and platforms have been proposed~\cite{frechet, guacamol, moses}. For example, Fréchet ChemNet Distance~\cite{frechet} is a metric that measures the distance between two distributions of molecules, generated and real, in terms of their chemical and biological properties. GuacaMol~\cite{guacamol} and Molecular sets (MOSES)~\cite{moses} are two other widely used benchmarking platforms that are composed of multiple metrics, such as validity, novelty, uniqueness, Fréchet ChemNet Distance, among others. These metrics assess the generated molecular output in terms of their validity, diversity, novelty, uniqueness, required physico-chemical properties, as well as their similarity to the training set. In other words, their focus is to provide quantitative and standardized descriptors to characterize the outputs of the generative methods. While they provide extensive characteristics, they do not evaluate the molecular output in a biological context. In particular, the ligand-target interaction process, which is of paramount importance to the drug discovery process, is currently not given sufficient attention in these benchmarks.

To this end, we propose to extend the current set of benchmark metrics with a number of tools that also take biological properties into account such as ligand-target interaction, which is named \emph{affinity binding and recreation applied heuristic for the analysis of molecules (ABRAHAM)}. The recreation of originator molecules (ROOM) is such a novel metric that we introduce. It aims at quantifying the likelihood of a given generative model to recreate known ligands for a specific protein target. After a set of ligands is assembled for a specific target, the model trains on a ligand subset and is tasked to generate an analogous output which is then compared to the true binders and their overlap is recorded. Next, we propose to incorporate machine learning (ML)-based drug-target affinity (DTA) binding predictors~\cite{rise_of_deep_learning, KronRLS, ozturk_deepdta_2018, Nguyen684662, DeepPurpose, multi-pli, GanDTI} for the evaluation of the generative models. In addition, we propose to apply molecular docking, which is the leading methodology of structure-based virtual screening~\cite{jastrzebski_emulating_2020}, to the outputs of the generative models in order to calculate ligand docking scores. These scores are subsequently utilized as an alternative and independent verification of the two above mentioned metrics.

To establish our proposed metrics we applied our pipeline to three important drug targets, namely vitamin D (VDR) and $\gamma$-aminobutyric acid type \textsubscript{A} (GABA\textsubscript{A}) receptors as well as the mammalian target of rapamycin (mTOR).

\begin{figure}[ht]
	\centering
	\captionsetup{type=figure}
	\includegraphics[width=\textwidth]{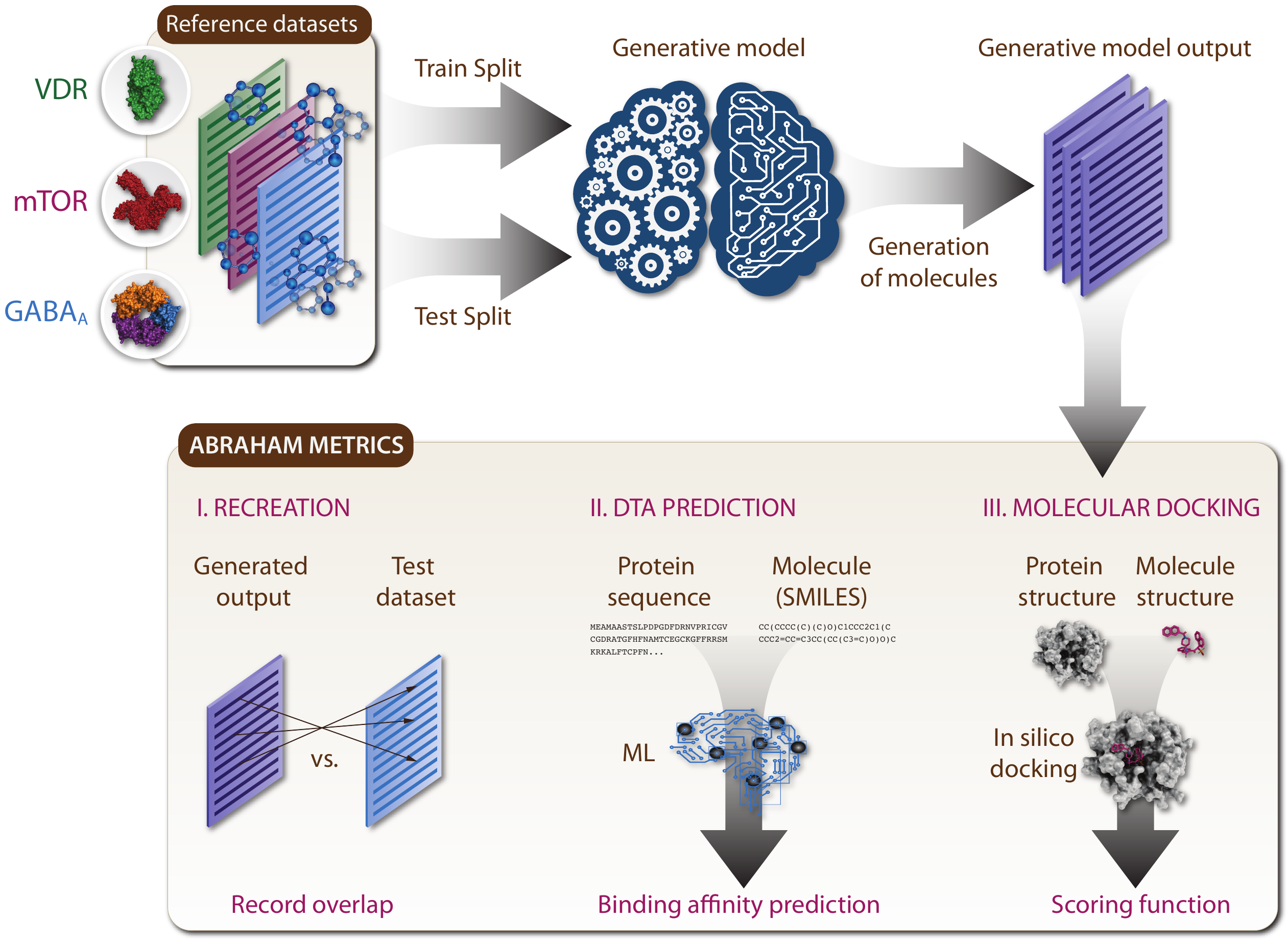}
	\caption{Graphical overview of the workflow for the ABRAHAM metrics.}
	\label{fig:overview}
\end{figure}

In short, this work introduces a novel biologically-inspired benchmark for evaluating \emph{de novo} molecular generation frameworks. A graphical representation of the workflow is depicted in Figure~\ref{fig:overview}.

\section{Results} \label{sec:Results}

\subsection{ROOM metric results}
\begin{figure}[ht]
	\centering
	\captionsetup{type=figure}
	\includegraphics[width=\textwidth]{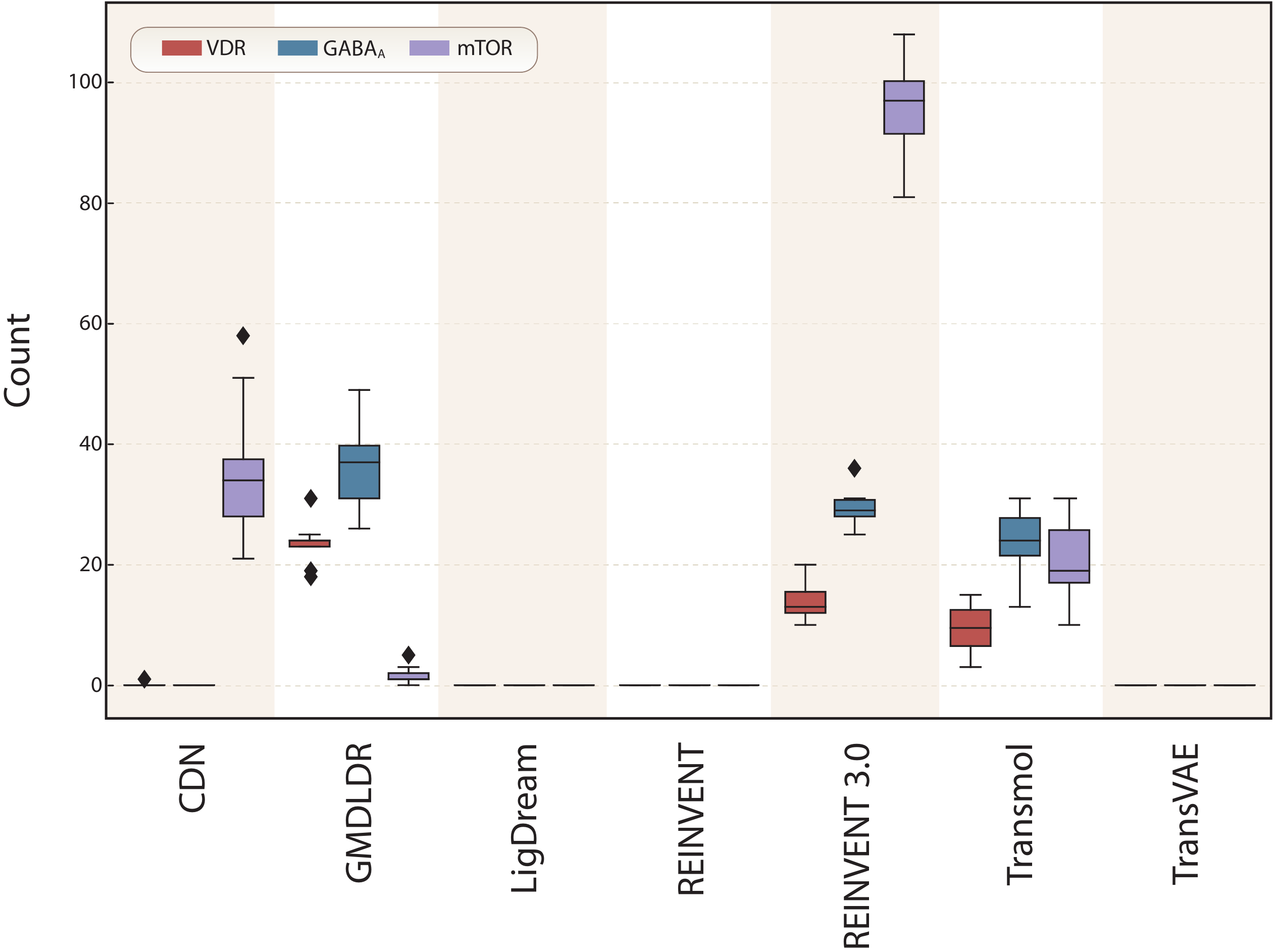}
	\caption{The average number of recreated ligands. The box-plots show the average number of recreated ligands for the 7 generative models across splits.}
	\label{fig:boxplot_generation_results}
\end{figure}

Table~\ref{tab:vdr-table} shows the overall results of the proposed ROOM recreation metric for all three reference datasets and the following seven generative models which have been used for this study: Conditional Diversity Network (CDN)~\cite{cdn_10.1145/3219819.3219882}; LigDream~\cite{ligdream_skalic2019shape}; Generative Molecular Design in Low Data Regime (GMDLDR)~\cite{Moret2020}; REINVENT~\cite{reinvent}; REINVENT 3.0~\cite{reinvent3}; Transmol~\cite{transmol_D1RA03086H} and TransVAE~\cite{TransVAE}.
In particular, the average number of recreated ligands across splits, the total number of recreated ligands (including the percentage of recreated ligands vs all test set ligands), the total number of unique generated molecules, the average number of generated molecules across splits and the ratio of recreated ligands by total number of generated molecules is shown. Values for best performing models are highlighted. Box-plots in Figure~\ref{fig:boxplot_generation_results} depict the average number of recreated ligands across splits.

\begin{table}[!ht]
	\caption{Recreation metrics for protein target datasets}
	\resizebox{\textwidth}{!}{
		\begin{tabular}{p{21mm}p{20mm}p{20mm}p{15mm}p{25mm}p{25mm}}
			\toprule
			             & average recreated ligands across 10 splits (Mean and std) & total \# unique recreated molecules & \# of valid unique generated molecules & average generated molecules across 10 splits (Mean and std) & ratio of recreated ligands to generated molecules ($10^{-6}$) \\  \midrule
			\multicolumn{6}{c}{{\bfseries VDR}}                                                                                                                                                                                                                                                   \\
			\midrule
			CDN          & 0.1±0.3                                                   & 1 (0.27\%)                          & 2468                                   & 257.8±101.9                                                 & 405                                                           \\
			GMDLDR       & 23.5±3.5                                                  & {\bfseries 100 (27\%)}              & 22458                                  & 2697.0±164.4                                                & 4453                                                          \\
			LigDream     & 0.0±0.0                                                   & 0                                   & 16717                                  & 2216.4±136.0                                                & 0                                                             \\
			REINVENT     & 0.0±0.0                                                   & 0                                   & 1176687                                & 118943.2±350.8                                              & 0                                                             \\
			REINVENT 3.0 & 13.9±3.2                                                  & 56 (15\%)                           & 6908                                   & 782.4±27.1                                                  & {\bfseries 8107}                                              \\
			Transmol     & 9.3±3.9                                                   & 63 (17\%)                           & 205969                                 & 15131.6±1,458.0                                             & 306                                                           \\
			TransVAE     & 0.0±0.0                                                   & 0                                   & 112512                                 & 12363.5±389.8                                               & 0                                                             \\
			\midrule
			\multicolumn{6}{c}{{\bfseries GABA\textsubscript{A}}}                                                                                                                                                                                                                                 \\
			\midrule
			CDN          & 0.0±0.0                                                   & 0                                   & 955                                    & 110.9±59.9                                                  & 0                                                             \\
			GMDLDR       & 35.9±6.8                                                  & 120 (47\%)                          & 10872                                  & 1498.2±108.9                                                & 11038                                                         \\
			LigDream     & 0.0±0.0                                                   & 0                                   & 20512                                  & 2790.9±135.9                                                & 0                                                             \\
			REINVENT     & 0.0±0.0                                                   & 0                                   & 1192523                                & 120735.4±186.3                                              & 0                                                             \\
			REINVENT 3.0 & 29.2±3.2                                                  & 110 (43\%)                          & 5632                                   & 731.6±20.8                                                  & {\bfseries 19531}                                             \\
			Transmol     & 24.0±5.2                                                  & {\bfseries 129 (51\%)}              & 168991                                 & 19245.6±3,779.7                                             & 767                                                           \\
			TransVAE     & 0.0±0.0                                                   & 0                                   & 415591                                 & 47894.9±379.7                                               & 0                                                             \\
			\midrule
			\multicolumn{6}{c}{{\bfseries mTOR}}                                                                                                                                                                                                                                                  \\
			\midrule
			CDN          & 35.6±11.3                                                 & 225 (4.86\%)                        & 20413                                  & 2238.6±220.7                                                & 11022                                                         \\
			GMDLDR       & 1.6±1.5                                                   & 13 (0.28\%)                         & 23271                                  & 2340.8±1,215.7                                              & 559                                                           \\
			LigDream     & 0.0±0.0                                                   & 0                                   & 453602                                 & 53509.9±517.0                                               & 0                                                             \\
			REINVENT     & 0.0±0.0                                                   & 0                                   & 1188355                                & 119769.3±365.5                                              & 0                                                             \\
			REINVENT 3.0 & 95.6±7.7                                                  & {\bfseries 645 (14.0\%)}            & 16947                                  & 1899.4±8.5                                                  & {\bfseries 38060}                                             \\
			Transmol     & 20.7±6.2                                                  & 177  (3.82\%)                       & 149068                                 & 15131.6±1,458.0                                             & 1187                                                          \\
			TransVAE     & 0.0±0.0                                                   & 0                                   & 111855                                 & 12354.9±151.0                                               & 0                                                             \\
			\bottomrule
		\end{tabular}
		\label{tab:vdr-table}
	}
\end{table}

According to the ROOM metric LigDream, REINVENT and TransVAE failed to recreate any test set ligands (see Table~\ref{tab:vdr-table}). While CDN showed inadequate results generating 1 and 0 ligands for VDR and GABA\textsubscript{A}, respectively, the performance is drastically improved for mTOR, where 225 ligands are recreated, which corresponds to a 4.86\% recreation rate of all test set ligands (see Table \ref{tab:vdr-table}).
GMDLDR recreates 100 (27\%), 120 (47\%) and 13 (0.28\%) ligands, for VDR, GABA\textsubscript{A} and mTOR, respectively.
REINVENT 3.0 performs well for all three targets. In terms of the ratio of recreated ligands by total number of generated molecules the model scores highest for each target.
Transmol is also able to recreate ligands for all datasets. In particular, for GABA\textsubscript{A} the recreation rate is 51\%, which is the highest among all seven generative models.

\begin{figure}[ht]
	\centering
	\captionsetup{type=figure}
	\includegraphics[width=\textwidth]{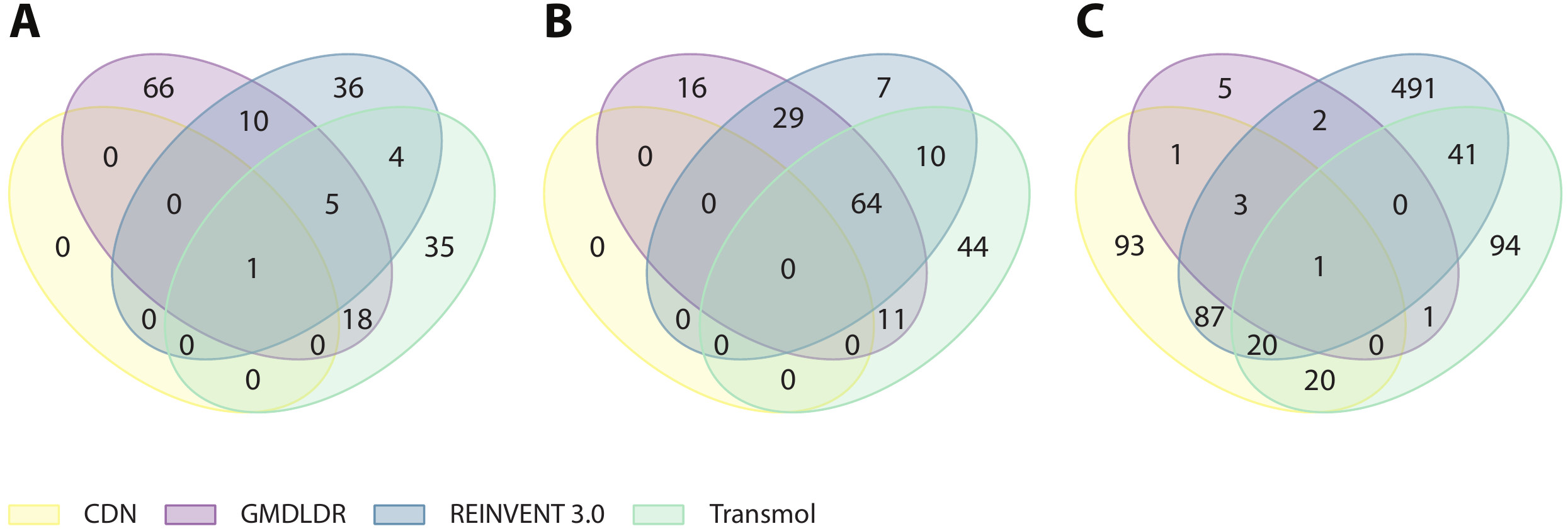}
	\caption{Overlap in the ligand recreation metrics. The Venn diagram shows the relationships among the recreated ligands, in particular the overlap for recreated ligands by the four best-performing generative models  CDN, GMDLDR, REINVENT 3.0 and Transmol, and three targets VDR (A) GABA\textsubscript{A} (B) and mTOR (C).}
	\label{fig:venn_generation_results_vdr}
\end{figure}

Figure~\ref{fig:venn_generation_results_vdr} shows three Venn diagrams for a) VDR, b) GABA\textsubscript{A} and c) mTOR datasets, which indicate the intersections of ligand recreation. Only those generative models are depicted that were able to recreate at least one test-set ligand, namely CDN, GMDLDR, REINVENT 3.0 and Transmol. While some ligands are jointly recreated by multiple generative models, others are uniquely recreated. For example in VDR, 137 ligands are recreated by a single generative model whereas 38 ligands are recreated simultaneously by at least two models. This indicates that the generative models complement each other for the ligand recreation process.

\subsection{DTA prediction metric results}

\begin{figure}[ht]
	\centering
	\captionsetup{type=figure}
	\includegraphics[width=0.9\textwidth]{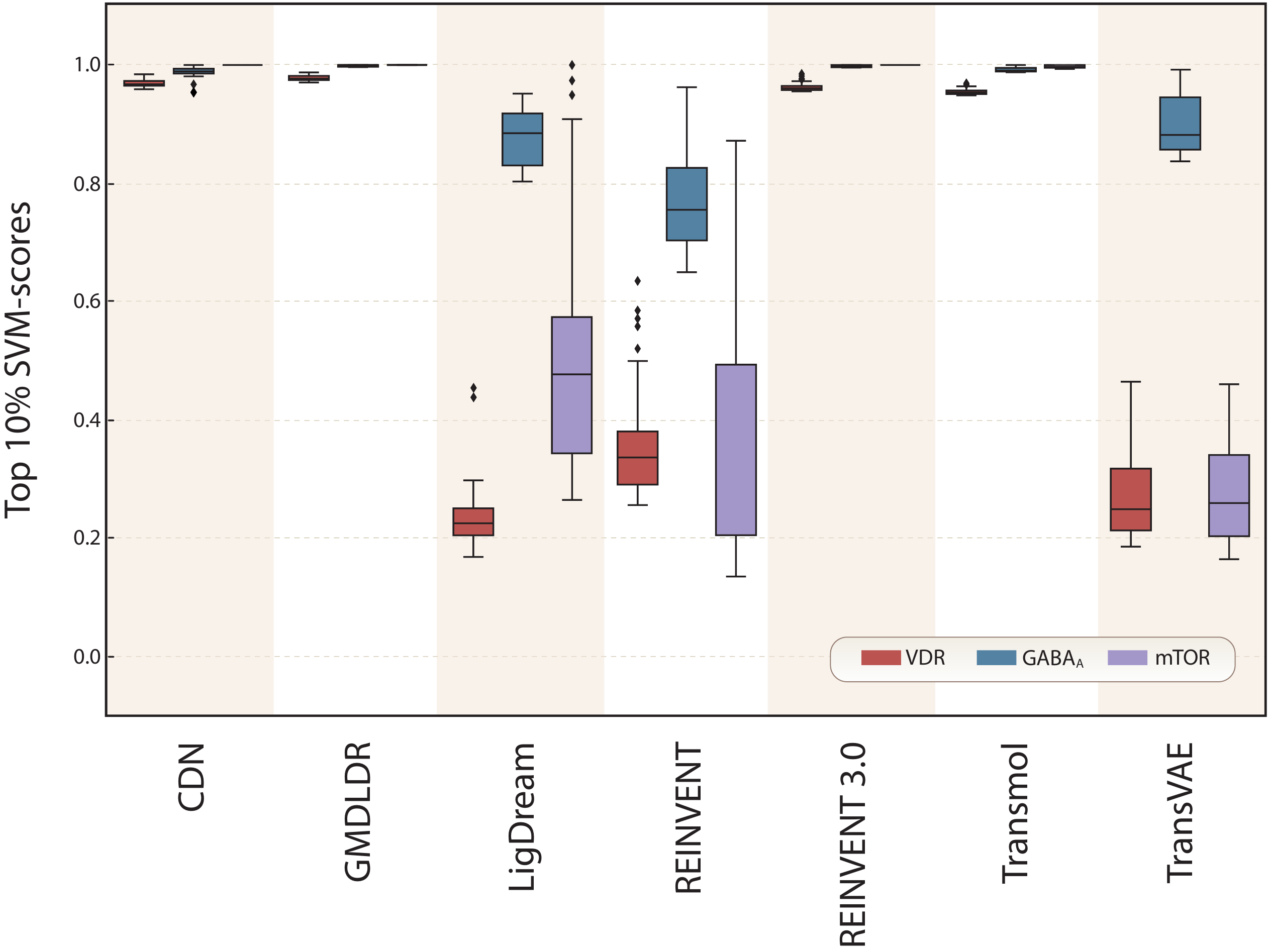}
	\caption{The SVM classification outputs. The box-plot shows the SVM classification outputs for the 7 generative models and three reference targets VDR, GABA\textsubscript{A} and mTOR.}
	\label{fig:svm_bar}
\end{figure}

Table~\ref{tab:tab_scores} gives a summary for the out-of-sample DTA predictions for the support vector machine (SVM), GraphDTA and multitask label encoding model (MLT-LE) models. The DTA predictions are initially performed for whole set of 400 molecules. Subsequently, predictions are sorted and only the top 10\% highest classification/regression outputs are shown. Additionally, box-plot representations of these SVM predictions are depicted in Figure~\ref{fig:svm_bar}. In Table~\ref{tab:tab_scores}, highest classification/regression outputs for a given model and reference dataset are highlighted in bold. Similar to the ROOM metric, also here CDN, GMDLDR, REVINVENT 3.0 as well as Transmol are the best performing models, i.e. their generative output is most likely to create ligands for the reference targets.

\begin{figure}[ht]
	\centering
	\captionsetup{type=figure}
	\includegraphics[width=\textwidth]{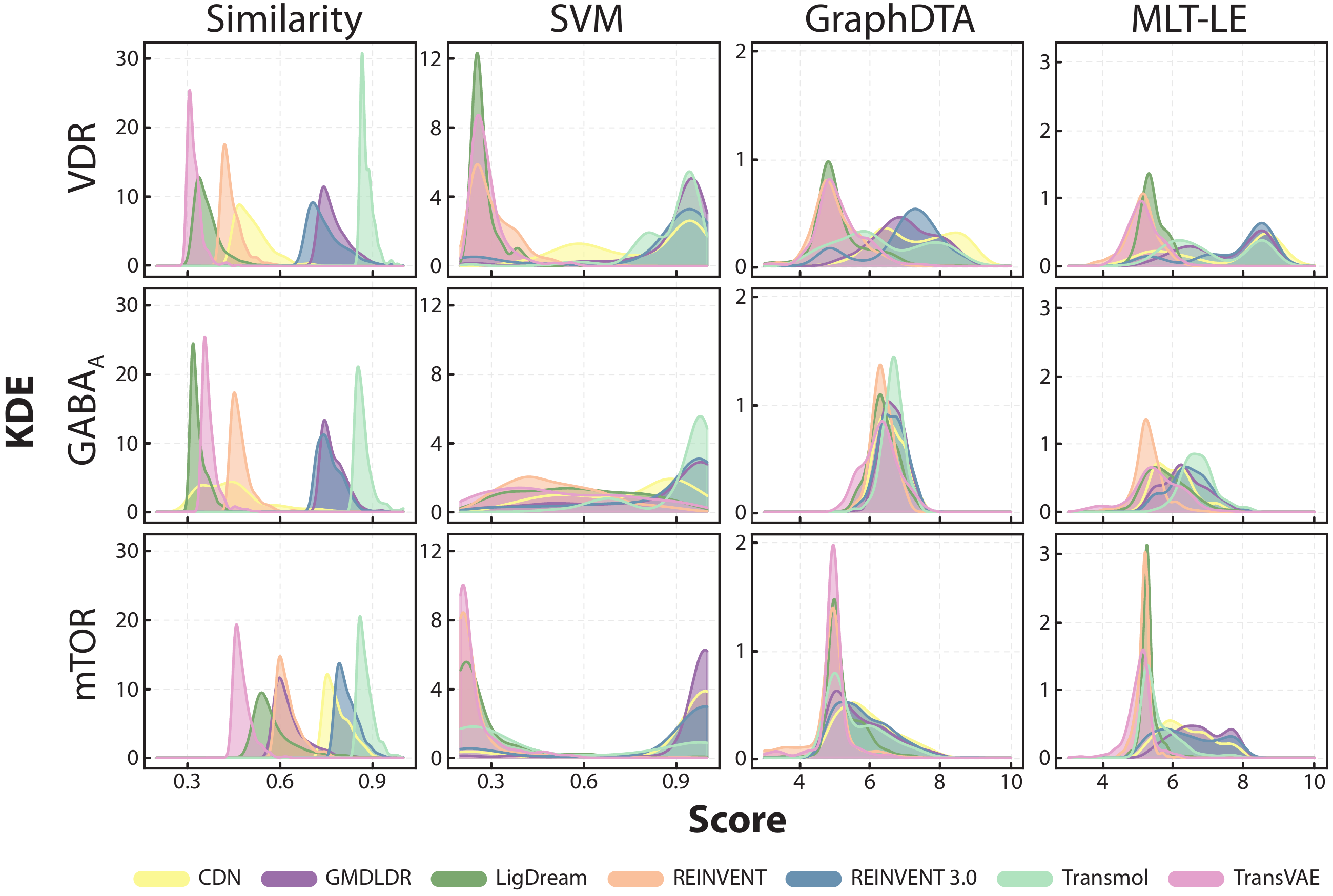}
	\caption{Distribution of the properties for the DTA-based prediction metrics. Kernel density estimation (KDE) of similarity between input and generative output (first column), SVM classification outputs (second column), GraphDTA (third column) and MLT-LE pKD predictions (fourth columns) for the 7 generative models and three targets VDR (first row), GABA\textsubscript{A} (second row) and mTOR (third row).}
	\label{fig:svm_sim}
\end{figure}

Figure~\ref{fig:svm_sim} shows kernel density estimates (KDE) for similarity, SVM, GraphDTA, and MLT-LE scores for all generative methods. KDE for similarity shows that Transmol creates focused libraries with high similarity to the reference set. GMDLDR and REINVENT 3.0 also show this type of behaviour, albeit to a lesser degree (left column of Figure~\ref{fig:svm_sim}). While KDE estimates of SVM outputs show bi-modal distributions across generative models, GraphDTA and MLT-LE exhibit mostly uni-modal distributions. 

\begin{figure}[ht]
	\centering
	\captionsetup{type=figure}
	\includegraphics[width=\textwidth]{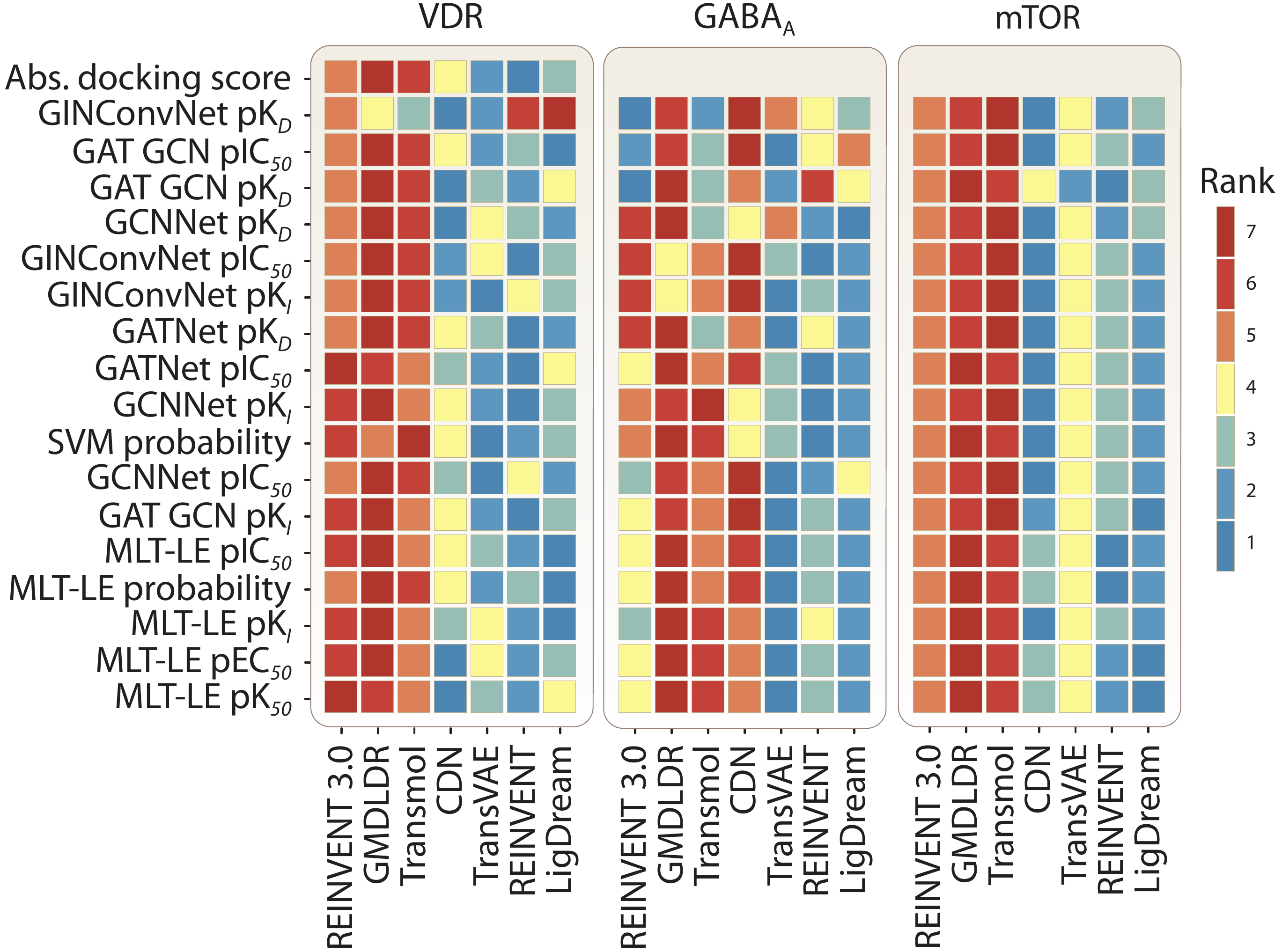}
	\caption{Ranking of generative models based on the DTA predictors and docking scores. The figure shows the ranking of the 7 generative models and three targets VDR GABA\textsubscript{A}and mTOR based on DTA predictors and docking scores. The ranking in dark red color (7) shows the best-performing generative models, while the blue color (1) the worst.}
	\label{fig:many_rankings1}
\end{figure}

In order to create a ranking of generative models (and molecular docking), 10\% of the output molecules with highest DTA predictions (respectively absolute docking scores for VDR) were selected and their predictions averaged. Generative models were then ranked according to the average prediction score of their generative output. Figure~\ref{fig:many_rankings1} shows the ranking of all considered generative methods. The highest rank corresponds to the highest average binding affinity, i.e. to high probability of ligand creation. As can be seen in Figure~\ref{fig:many_rankings1} ranking across generative models is mostly conserved for mTOR, while more variability in the ranking across methods is observed for VDR and more so for GABA\textsubscript{A}. The top row shows the ranking according to the docking scores for VDR. Interestingly, the ranking concurs with the DTA prediction results.

\begin{figure}[ht]
	\centering
	\captionsetup{type=figure}
	\includegraphics[width=\textwidth]{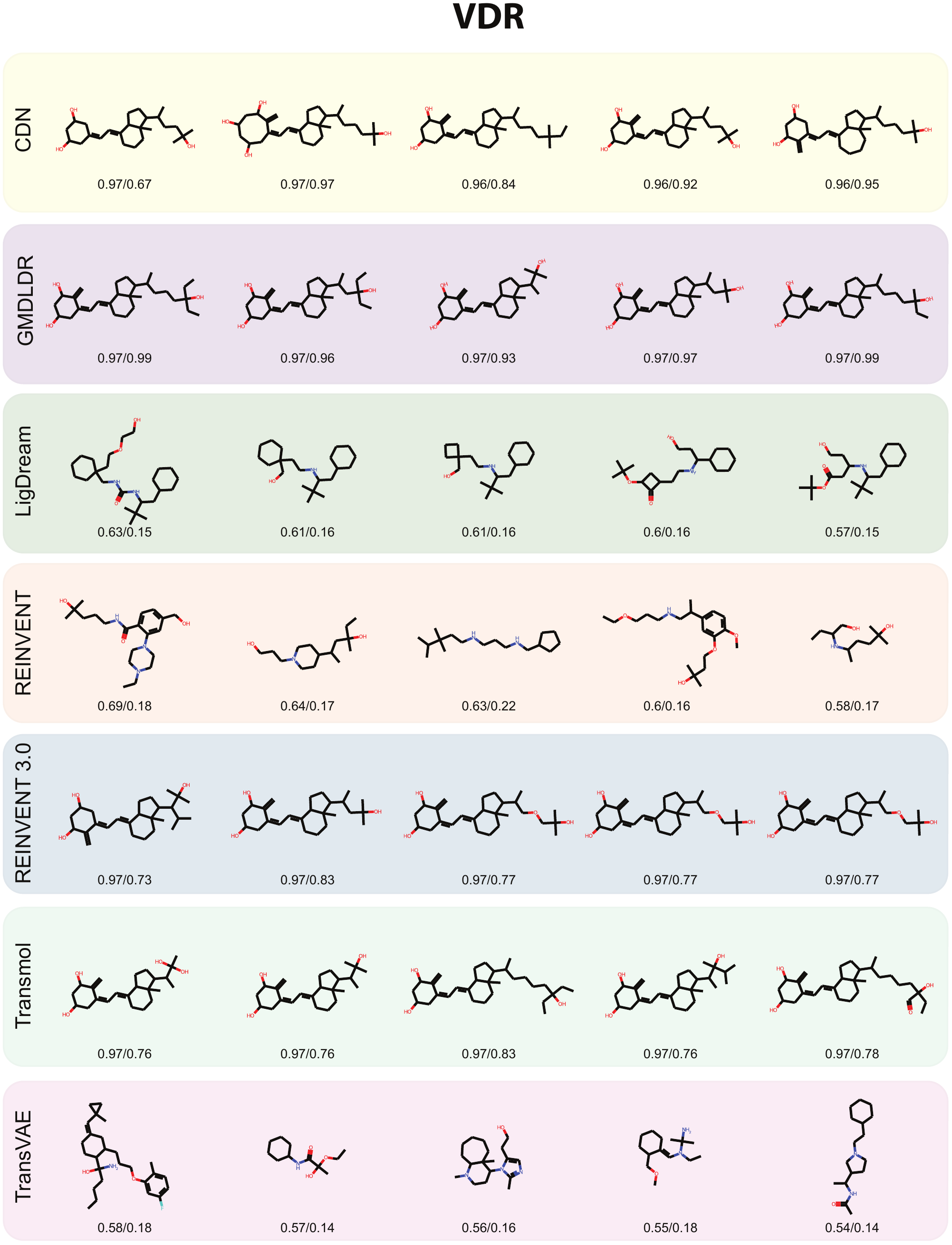}
	\caption{Top 5 generated molecules sorted by SVM output for VDR. The two values under each structure represent the SVM and similarity scores, respectively.}
	\label{fig:svm_top5_vdr}
\end{figure}

\begin{figure}[ht]
	\centering
	\captionsetup{type=figure}
	\includegraphics[width=\textwidth]{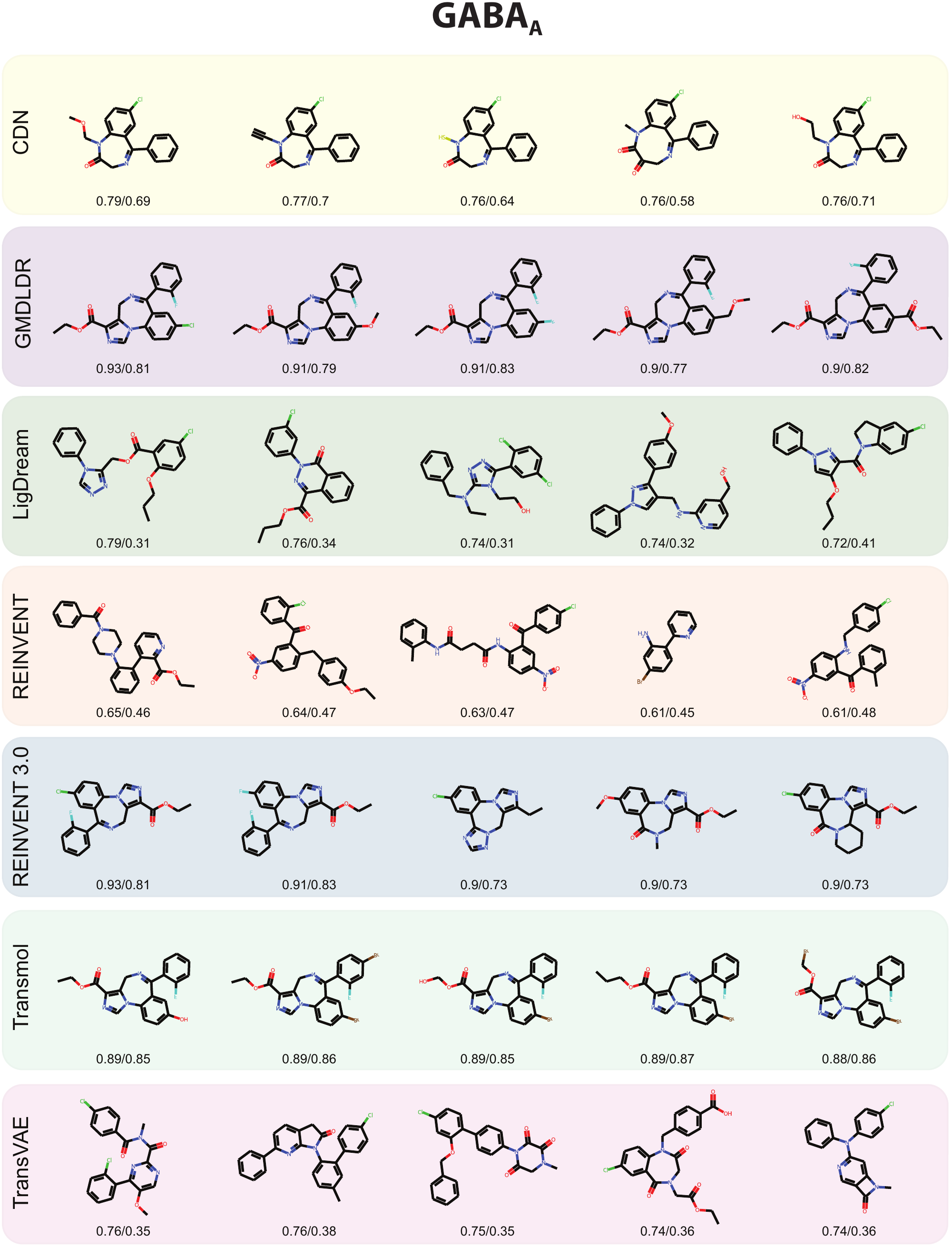}
	\caption{Top 5 generated molecules sorted by SVM output for GABA\textsubscript{A}. The two values under each structure represent the SVM and similarity scores, respectively.}
	\label{fig:svm_top5_gaba}
\end{figure}

\begin{figure}[ht]
	\centering
	\captionsetup{type=figure}
	\includegraphics[width=\textwidth]{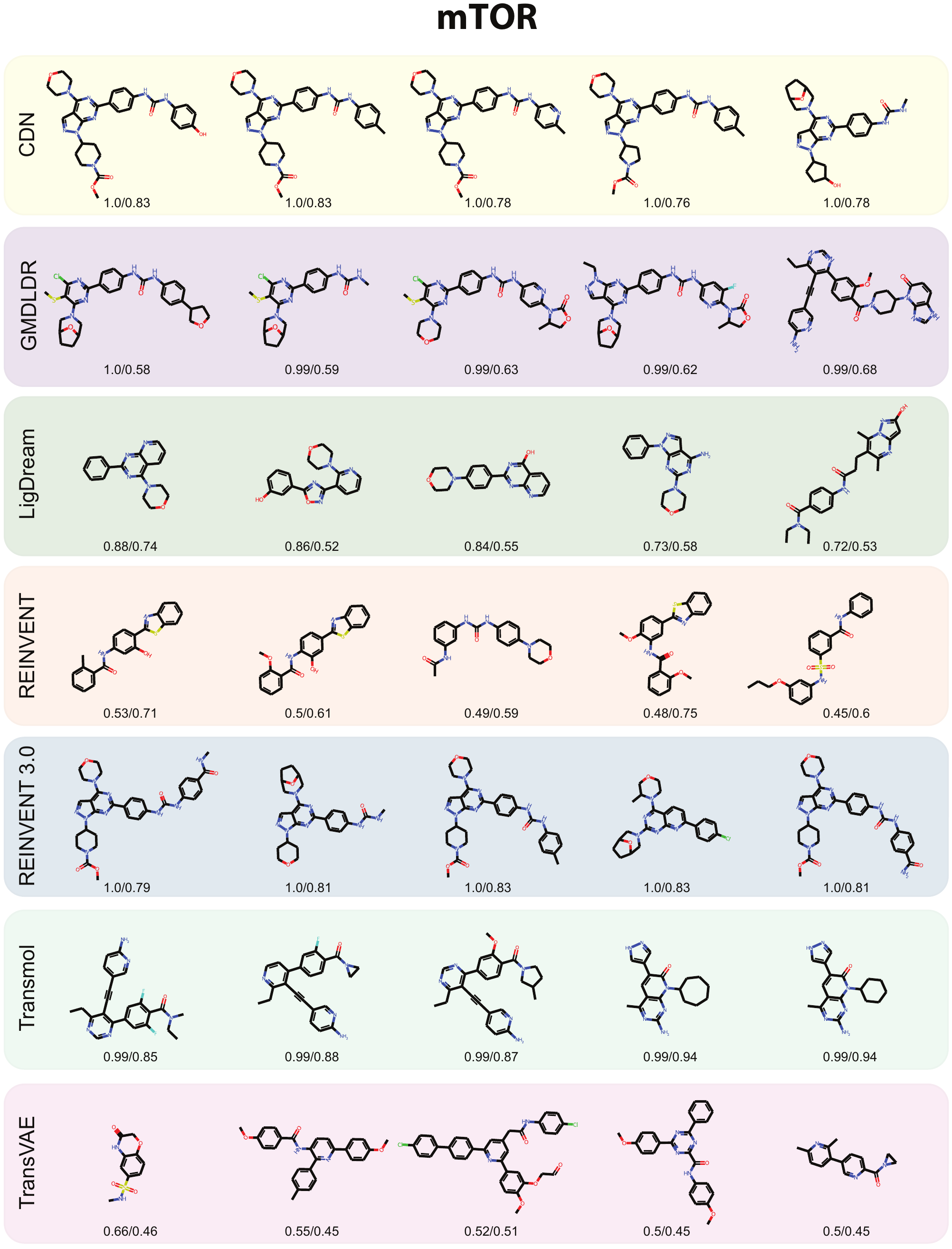}
	\caption{Top 5 generated molecules sorted by SVM output for mTOR. The two values under each structure represent the SVM and similarity scores, respectively.}
	\label{fig:svm_top5_mtor}
\end{figure}

Figures~{\ref{fig:svm_top5_vdr}},~{\ref{fig:svm_top5_gaba}} and~{\ref{fig:svm_top5_mtor}} show the top five structures sorted by their SVM DTA prediction score for each generative method and target. For VDR, a detailed inspection of the top five molecular structures generated by CDN, GMDLDR, REINVENT 3.0 and Transmol revealed that all of them contain the key structural features of VDR binding molecules of the secosteroidal type. For GABA\textsubscript{A}, four out of the top five molecules generated by CDN exhibit classical benzodiazepine (BZD) scaffold, consisting of a fused benzene and diazepine ring, while one molecule featured an expanded benzodiazocine core. Interestingly, all the top molecules generated by GMDLDR, REINVENT 3.0 and Transmol contain the polycyclic benzo[f]imidazo[1,5-a][1,4]diazepine scaffold in which the benzodiazepine moiety is fused with an additional imidazole ring. Finally, one molecule containing the BZD scaffold is found among the top five structures produced by TransVAE. Due to the large variability of the mTOR ligands, unique scaffolds or structural fragments are more difficult to identify. Furthermore, while the overlap of scaffolds between molecules across the outputs of different generative models does occur, it is not as extensive as for VDR and GABA\textsubscript{A}. In particular, the most frequently occurring 1-(4-(1H-pyrazolo[3,4-d]pyrimidin-6-yl)phenyl)urea fragment appears in five, four and one of the top five molecules generated by CDN, REINVENT 3.0 and GMDLDR, respectively. In addition, a structurally related 1-(4-(pyrimidin-2-yl)phenyl)urea fragment is found in three out of the top five molecules produced by GMDLDR. Another frequently found fragment, 5-((4-phenylpyrimidin-5-yl)ethynyl)pyridin-2-amine, appears in two and one of the top five molecules stemming from Transmol and GMDLDR, respectively.
It should be noted that different scaffold distributions can be obtained if the top five molecules are selected based on Tanimoto similarity score or on binding strength indicators as opposed to SVM score. The respective molecular structures produced by each generative model for all three reference targets can be found in the supplementary material.

\begin{table}[!ht]
	\caption{The mean value of the highest 10\% of the scores for each generative method}
	\resizebox{\textwidth}{!}{
		\begin{tabular}{cccccccccc}
			\toprule
			DTA method    & \multicolumn{3}{l}{SVM (is $\mu M < 1$)} & \multicolumn{3}{l}{GCNNet (*pK\textit{\textsubscript{D}})} & \multicolumn{3}{l}{MLT-LE (*pK\textit{\textsubscript{D}})}                                                                                                                                 \\ \midrule
			              & VDR                                      & GABA\textsubscript{A}                                      & mTOR                                                       & VDR               & GABA\textsubscript{A} & mTOR              & VDR               & GABA\textsubscript{A} & mTOR              \\ \midrule
			CDN           & 0.947                                    & 0.979                                                      & 0.998                                                      & {\bfseries 8.901} & 7.285                 & {\bfseries 7.479} & {\bfseries 8.728} & 7.087                 & 7.70              \\
			GMDLDR        & {\bfseries 0.961}                        & {\bfseries 0.984}                                          & {\bfseries 0.999}                                          & 8.447             & 7.284                 & 7.341             & 8.692             & 7.385                 & 7.788             \\
			LigDream      & 0.272                                    & 0.902                                                      & 0.533                                                      & 6.634             & 7.09                  & 6.323             & 6.068             & 6.971                 & 5.784             \\
			REINVENT      & 0.355                                    & 0.749                                                      & 0.449                                                      & 6.29              & 6.923                 & 5.914             & 5.89              & 6.055                 & 5.434             \\
			REINVENT  3.0 & 0.942                                    & 0.983                                                      & {\bfseries 0.999}                                          & 8.44              & {\bfseries 7.299}     & 7.338             & 8.692             & 7.539                 & {\bfseries 7.793} \\
			Transmol      & 0.936                                    & 0.981                                                      & 0.986                                                      & 8.656             & 7.173                 & 7.143             & 8.656             & {\bfseries 7.667}     & 6.801             \\
			TransVAE      & 0.278                                    & 0.881                                                      & 0.432                                                      & 6.446             & 7.222                 & 5.677             & 6.043             & 6.639                 & 5.643             \\ \midrule
			\multicolumn{10}{l}{\textit{* $pK\textit{\textsubscript{D}} = -log_{10}(\frac{K\textit{\textsubscript{D}}}{10^9} + 1e-10)$}}                                                                                                                                                                                       \\ \bottomrule
		\end{tabular}
		\label{tab:tab_scores}
	}
\end{table}

\subsection{Molecular docking results}
\begin{figure}[ht]
	\centering
	\captionsetup{type=figure}
	\includegraphics[width=\textwidth]{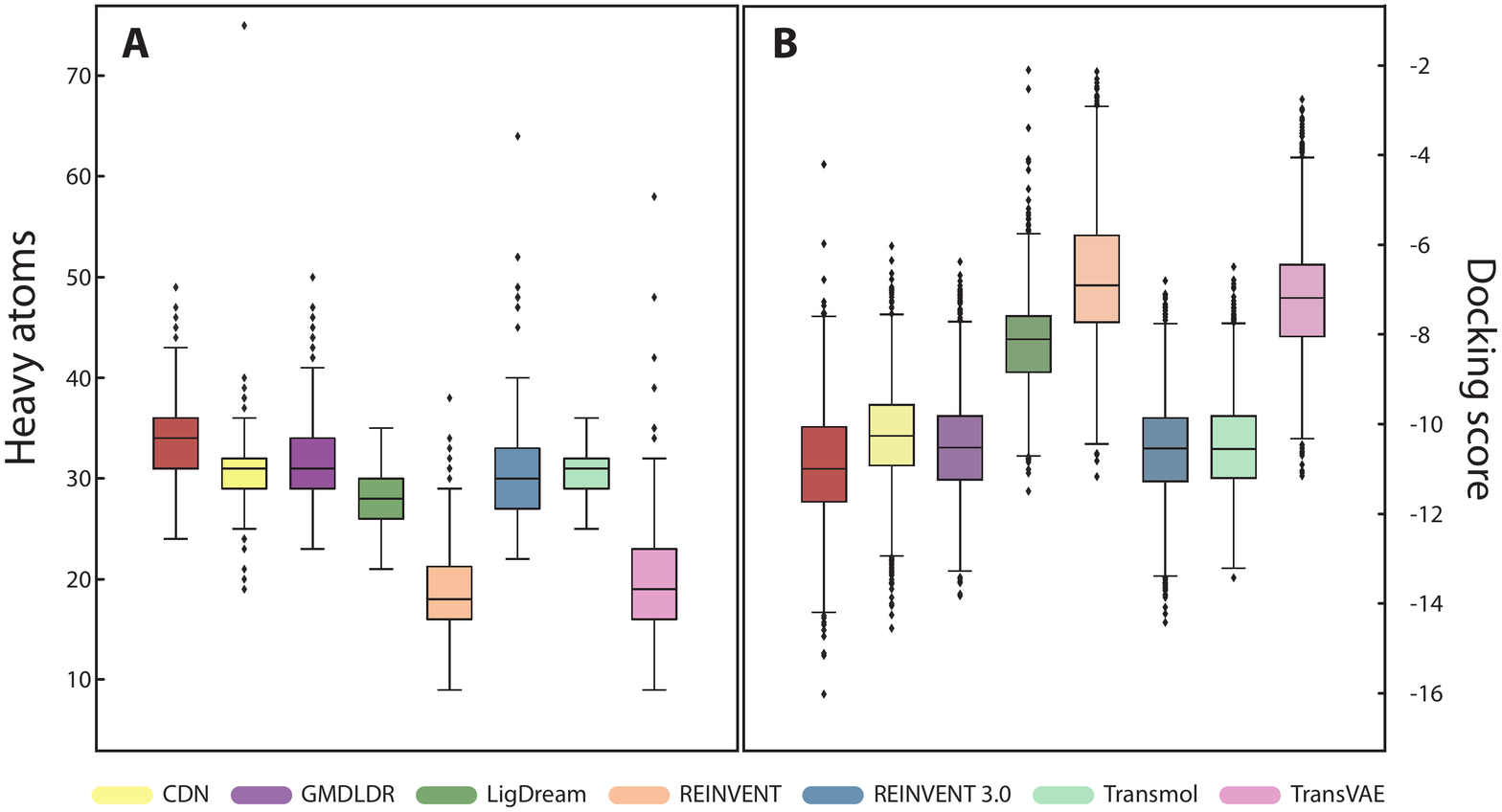}
	\caption{Distribution of ligand heavy atom counts and docking scores. The box-plots show the heavy atom count (A) and docking scores (B) for the VDR reference dataset (in red) and the outputs of 7 generative models. Lower docking scores mean higher binding affinities.}
	\label{fig:docking_dist1}
\end{figure}

A comparison of heavy atom count between the VDR reference and generated molecules revealed similar distributions for five of the seven generated output datasets. REINVENT and TransVAE had smaller average heavy atom counts compared to the reference dataset and other generative models (Figure~\ref{fig:docking_dist1}). In addition, the generated molecules from REINVENT and TransVAE had lower average heavy atom count as compared to the reference dataset. Based on this analysis and the low number of identified outliers, the Glide SP molecular docking was performed without further filtering.

The validation of the Glide SP docking procedure using the top-5 poses confirmed the feasibility of this approach. In particular, the conformation of the docked molecule was in agreement with their orientation compared to the original PDB structures or calcitriol in case of C2$\alpha$-butyl-calcitriol analog (see supplementary data).

\begin{figure}[ht]
	\centering
	\captionsetup{type=figure}
	\includegraphics[width=\textwidth]{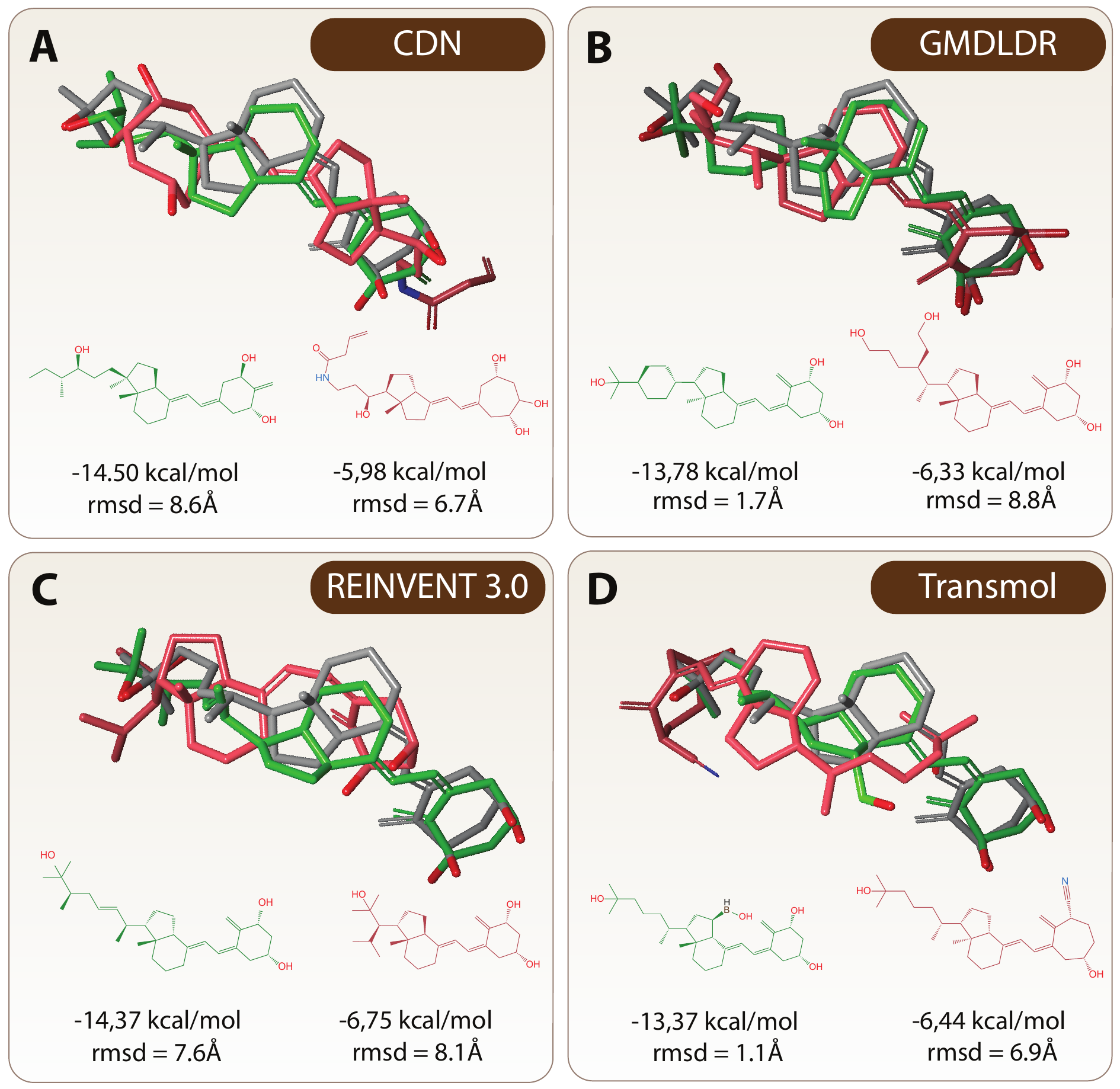}
	\caption{Illustrative docking results for the four best-performing generative models. Comparison of the best (green) and the worst (red) poses for the top compounds generated by the four best-performing algorithms CDN (A), GMDLDR (B), REINVENT 3.0 (C) and Transmol (D) to the reference calcitriol with its best re-docking pose (gray). The chemical structure of the ligands is depicted under their three-dimensional representation. The RMSD was calculated in-place against the reference (-12.03 kcal/mol, rmsd = 1.31\text{\AA}).}
	\label{fig:docking_4x4}
\end{figure}

The results of molecular docking showed that ligands from four generative models, namely CDN, GMDLDR, REINVENT 3.0 and Transmol have their distribution of docking scores in the same range as the VDR reference dataset (Figure~\ref{fig:docking_dist1}, Table~\ref{tab:docking_table1}). Furthermore, superimposition based on the most common substructure of the "best" (Figure~\ref{fig:docking_4x4}A-D, highlighted in green) and the "worst" (Figure~\ref{fig:docking_4x4}A-D, highlighted in red) pose, from the top performing generative models, to the best calcitriol pose, from the VDR reference dataset, was further explored and visualized (Figure~\ref{fig:docking_4x4}). Root-mean square distance (RMSD), the measure of similarity for the spatial conformation, indicates the lowest value 1.1 {\it \r{A}} for Transmol vs. 1.31 {\it \r{A}} for the best pose of calcitriol (Figure~\ref{fig:docking_4x4}D). The calculated binding energies for the best poses are in the range of -14 kcal/mol. In contrast, the best pose of calcitriol has a binding energy of -12.03 kcal/mol. For the four generative models, the best poses show a lower docking score than the best pose for calcitriol, which indicates a more favorable conformation and higher affinity. With respect to the "worst" poses the position of the CD-ring is significantly shifted out of any alignment and in some cases such as for Transmol the spatial orientation of the molecule is reversed as compared to calcitriol (Figure~\ref{fig:docking_4x4}D).

\begin{table}[!ht]
	\caption{Docking score statistics for VDR reference and generative  datasets}
	\resizebox{\textwidth}{!}{
		\begin{tabular}{ccccccc}
			\toprule
			              & mean ± std                & min               & 25\%               & 50\%               & 75\%               & max               \\ \midrule
			VDR reference & -10.87 ± 1.27             & -15.97            & -11.68             & -10.94             & -10.01             & -4.16             \\
			CDN           & -10.20 ± 1.05             & {\bfseries -14.5} & -10.87             & -10.21             & -9.52              & -5.98             \\
			GMDLDR        & -10.45 ± 1.06             & -13.78            & -11.19             & -10.47             & -9.77              & -6.33             \\
			LigDream      & -8.14 ± 1.00              & -11.44            & -8.79              & -8.06              & -7.54              & -2.05             \\
			REINVENT      & -6.70 ± 1.53              & -11.12            & -7.68              & -6.86              & -5.74              & -2.09             \\
			REINVENT 3.0  & {\bfseries -10.52 ± 1.07} & -14.39            & {\bfseries -11.22} & -10.49             & {\bfseries  -9.81} & {\bfseries -6.75} \\
			Transmol      & -10.43 ± 1.02             & -13.37            & -11.15             & {\bfseries -10.51} & -9.77              & -6.44             \\
			TransVAE      & -7.14 ± 1.25              & -11.1             & -7.99              & -7.14              & -6.4               & -2.71             \\ \bottomrule
		\end{tabular}
		\label{tab:docking_table1}
	}
\end{table}

\subsection{Divergence between DTA-based predictions and molecular docking}

\begin{figure}[ht]
	\centering
	\captionsetup{type=figure}
	\includegraphics[width=\textwidth]{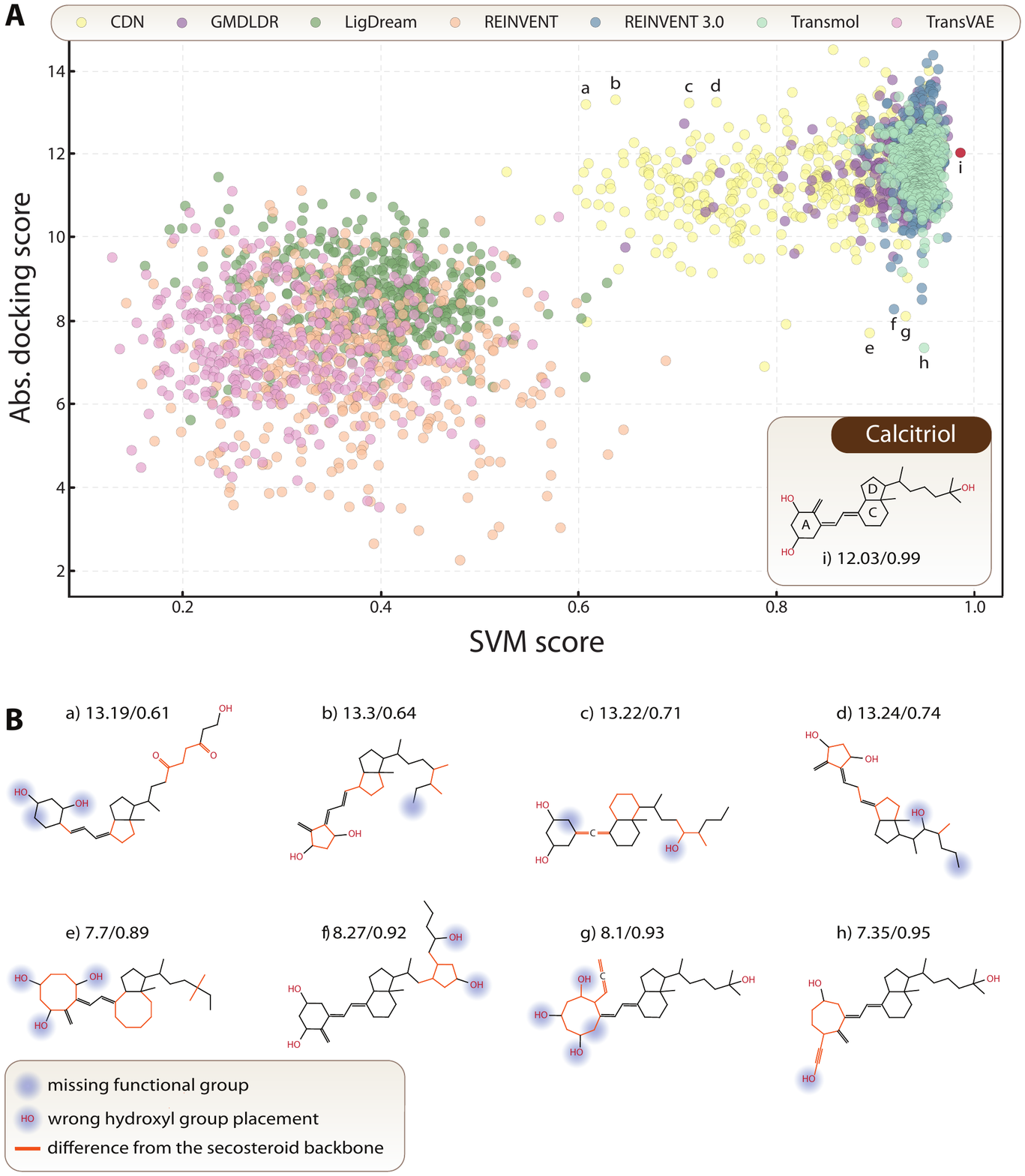}
	\caption{Comparison of generated ligands, based DTA prediction and molecular docking scores. The scatter plot depicts molecular outputs from all 7 generative models and compares SVM (x-axis) and absolute docking scores (y-axis). The outputs from generative models are color-coded. For illustration, calcitriol is shown as a red circle (molecule 'i') (A). Illustrative molecules with high docking, but low SVM scores ('a-d'; first row) and molecules with low docking, but high SVM scores ('e-h'; second row) are depicted. Their deviations from reference calcitriol are highlighted by orange color and the missing functional or incorrectly positioned hydroxyl group by blue circles (B).}
	\label{fig:div}
\end{figure}

Calcitriol (see Figure~\ref{fig:div}A molecule `i'), bottom-right panel) is regarded as the most active natural metabolite of vitamin D\textsubscript{3} and contains a secosteroid backbone consisting of A-, C- and D-rings. The residual B-ring from the sterane molecule manifests in the form of a conjugated diene linker, having two exocyclic carbon atoms and connecting the A- to the C/D-rings, and a methylene group on the A-ring (see Figure~\ref{fig:div}A molecule `i'). The carefully positioned two hydroxyl (OH) groups on the A-ring at carbons C1 and C3 as well as the third OH at C25 at the very end of an aliphatic chain fused onto the D-ring have co-evolved with VDR for optimal receptor activation and subsequent regulation of target genes~\cite{Ferdi2012}.

Figure~\ref{fig:div}A depicts a scatter plot, where SVM-based DTA scores (x-axis) are plotted against absolute docking scores (y-axis) for all VDR generated molecules. The molecular outputs of the 7 generative models are color-coded. For reference, these scores are also plotted for calcitriol and highlighted as a red circle (Figure~\ref{fig:div}A compound `i'). Overall, DTA-based prediction scores correlate well with docking scores of potential VDR ligands (concordance index $= -0.77$). Individual molecules are distributed in two large clusters: a cluster of unlikely binders having low docking and DTA scores; and a cluster of likely binders for which both scores are relatively high indicating that the two scores correlate well for the majority of generated molecules.

However, there are also several scarcely populated regions where the two scores diverge. In particular, we were interested in examining two cases, where molecules exhibit a high docking score but a low DTA-based prediction score and vice versa. Close structural evaluation of molecules belonging to both of these regions identified molecules that on first sight seem to resemble the ideal secosteroid backbone inherent to calcitriol, yet they represent chimeras that differ from the classical secosteroid structure in various details. To illustrate this, the deviation from the secosteroid backbone in these compounds is highlighted in orange color and shown in Figure~\ref{fig:div} panel B.

In the first row of panel B the four molecules exhibit high docking but low SVM scores (Figure~\ref{fig:div}B molecules 'a-d'). Interestingly, all these molecules deviate from the classical secosteroid backbone in both, the ring system as well as their aliphatic chains. The aliphatic chains are usually branched and have a missing or incorrectly positioned OH group. For molecules `b' and `d', the C-rings are smaller while the linker between the two ring systems is longer compared to calcitriol. For molecule `c', the D-ring is enlarged while the linker is shortened. All these deviations may have led to the DTA-based prediction assigning lower scores. With respect to molecular docking results, all molecules were compared to the VDR-calcitriol crystal structure (PBDID 1DB1), where calcitriol has a ``boat-shaped'' conformation in the ligand-binding pocket (LBP) of VDR. Only `b' and `d' showed comparable conformation to calcitriol and the correct orientation in the LBP. While their correct conformation may have led to high docking scores, it should be noted that the position of OH groups as well as the orientation of these molecules are incorrect. As described above the increase in the length and/or size of the molecules due to a longer linker and branched aliphatic chain did not prevent the molecules to take the correct boat-shaped conformation. Molecule 'c' represents an interesting analog that is very similar to calcitriol in terms of its size and although docked in an incorrect orientation some of the OH group were positioned rather similarly compared to calcitriol.
Molecule 'a' has the longest aliphatic chain from all the four compounds (about 3.6 {\it \r{A}}), nonetheless the size of the LBP can easily accommodate it. Even though the orientation compared to calcitriol is reversed, it fully follows the boat-shaped confirmation.

The second row contains examples of molecules that show low docking but high SVM-scores (Figure~\ref{fig:div}B molecules `e-h'). From the four molecules that we examine here, `e' and `g' were generated by CDN. Both exhibit the incorporation of larger rings. The tendency of CDN to incorporate larger rings into the generated molecules can also be observed for two of the top five molecules with the highest SVM-based DTA prediction scores (see Figure~{\ref{fig:svm_top5_vdr}}).

Molecules `e', `g' and `h' show incorrect orientation as well as some additional steric issues in their conformations. For `e', which has a shorter aliphatic chain than calcitriol, all the ring sub-structures are shifted away from the residue W286, which creates a critical hydrophobic interaction with the C/D-rings of calcitriol. As expected, a very similar feature can be observed for molecule `g'. The most notable structural misalignment can be noticed for `h', which has the lowest docking score from all four molecules. With compensation for the long linker that contains one of the OH groups the molecule is shifted more than 3.5 {\it \r{A}} leading to a deformed boat-shaped confirmation. Although molecule `f' shows rather high structural conservation with calcitriol and a correct orientation in the LBP, the docking algorithm tried to compensate for the additional ring and length of the aliphatic chain by displacing the ring structures leading to a shift in the position of the molecule. The alignment of OH groups is also incorrect, which may have also contributed to the low docking score. For the limited examples that have been examined here, it may seem that DTA predictions assign high scores to molecules with a structurally similar backbone and overall size, whereas a high docking score can be expected for molecules that follow calcitriol's evolutionary conserved boat-shaped conformation and positioning OH groups. Thus in order for an individual ML-generated molecule to display appropriate structural features inherent of likely binders, both SVM-based DTA prediction and docking scores should be high. In addition to these scores, we believe that the manual inspection of the docking results reinforces the judgement on the correct conformation of the generated molecule for the respective protein target.

\subsection{MOSES metrics}

In addition to ABRAHAM, MOSES metrics were calculated and are given in Table~\ref{tab:moses-vdr}. Most of the generative models achieve high scores of around 0.8-1.0 with standard metrics such as validity, uniqueness, novelty, etc. Moreover, models that have good ABRAHAM scores, i.e. CDN, GMDLDR, REINVENT 3.0, and Transmol, performed well for MOSES metrics that measure similarities between the generated and testing sets of molecules. These metrics include fragment similarity (Frag/Test), scaffold similarity (Scaf/Test), similarity to a nearest neighbour (SNN/Test), and Fréchet ChemNet distance (FCD/Test).

\begin{table}[!ht]
	\centering
	\caption{MOSES metrics for generated datasets: VDR; GABA\textsubscript{A}; and mTOR}
	\resizebox{\textwidth}{!}{
		\begin{tabular}{@{}lllllll@{}}
			\toprule
			\multicolumn{6}{c}{{\bfseries VDR}}                                                                                                                              \\ \midrule
			\textbf{Method} & \textbf{valid}             & \textbf{unique@1000}       & \textbf{Filters}           & \textbf{Novelty}           & \textbf{IntDiv}            \\ \midrule

			CDN             & 0.922 $\pm$ 0.059          & 0.783 $\pm$ 0.091          & 0.737 $\pm$ 0.090          & \textbf{1.000 $\pm$ 0.000} & 0.761 $\pm$ 0.048          \\
			GMDLDR          & \textbf{1.000 $\pm$ 0.000} & 0.970 $\pm$ 0.011          & 0.939 $\pm$ 0.020          & 0.982 $\pm$ 0.002          & 0.763 $\pm$ 0.013          \\
			Ligdream        & \textbf{1.000 $\pm$ 0.000} & 0.939 $\pm$ 0.011          & 0.955 $\pm$ 0.008          & \textbf{1.000 $\pm$ 0.000} & 0.817 $\pm$ 0.001          \\
			REINVENT        & 0.970 $\pm$ 0.003          & \textbf{1.000 $\pm$ 0.001} & 0.830 $\pm$ 0.011          & \textbf{1.000 $\pm$ 0.000} & 0.842 $\pm$ 0.002          \\
			REINVENT 3.0    & 0.996 $\pm$ 0.002          & 0.786 $\pm$ 0.026          & \textbf{0.969 $\pm$ 0.006} & 0.952 $\pm$ 0.004          & 0.747 $\pm$ 0.018          \\
			TransVAE        & 0.600 $\pm$ 0.006          & \textbf{1.000 $\pm$ 0.000} & 0.925 $\pm$ 0.004          & \textbf{1.000 $\pm$ 0.000} & \textbf{0.865 $\pm$ 0.001} \\
			Transmol        & \textbf{1.000 $\pm$ 0.000} & 0.827 $\pm$ 0.028          & 0.797 $\pm$ 0.029          & 0.999 $\pm$ 0.000          & 0.816 $\pm$ 0.018          \\  \midrule

			                & \textbf{FCD/Test}          & \textbf{SNN/Test}          & \textbf{Frag/Test}         & \textbf{Scaf/Test}         & \textbf{IntDiv2}           \\ \midrule
			CDN             & 17.615 $\pm$ 2.737         & 0.376 $\pm$ 0.053          & 0.539 $\pm$ 0.096          & 0.123 $\pm$ 0.168          & 0.713 $\pm$ 0.048          \\
			GMDLDR          & \textbf{5.518 $\pm$ 0.566} & \textbf{0.597 $\pm$ 0.011} & \textbf{0.941 $\pm$ 0.024} & \textbf{0.865 $\pm$ 0.033} & 0.723 $\pm$ 0.011          \\
			Ligdream        & 36.024 $\pm$ 0.993         & 0.231 $\pm$ 0.002          & 0.472 $\pm$ 0.015          & 0.006 $\pm$ 0.005          & 0.804 $\pm$ 0.001          \\
			REINVENT        & 45.934 $\pm$ 1.589         & 0.223 $\pm$ 0.004          & 0.473 $\pm$ 0.028          & 0.060 $\pm$ 0.008          & 0.835 $\pm$ 0.002          \\
			REINVENT 3.0    & 10.717 $\pm$ 0.851         & 0.579 $\pm$ 0.012          & 0.763 $\pm$ 0.031          & 0.441 $\pm$ 0.041          & 0.709 $\pm$ 0.016          \\
			TransVAE        & 48.917 $\pm$ 0.950         & 0.200 $\pm$ 0.001          & 0.422 $\pm$ 0.018          & 0.020 $\pm$ 0.009          & \textbf{0.859 $\pm$ 0.001} \\
			Transmol        & 10.021 $\pm$ 0.960         & 0.531 $\pm$ 0.016          & 0.862 $\pm$ 0.050          & 0.735 $\pm$ 0.087          & 0.779 $\pm$ 0.016          \\ \midrule
			\multicolumn{6}{c}{{\bfseries GABA\textsubscript{A}}}                                                                                                            \\ \midrule
			\textbf{Method} & \textbf{valid}             & \textbf{unique@1000}       & \textbf{Filters}           & \textbf{Novelty}           & \textbf{IntDiv}            \\ \midrule

			CDN             & 0.880 $\pm$ 0.106          & 0.419 $\pm$ 0.196          & 0.440 $\pm$ 0.245          & 0.996 $\pm$ 0.005          & 0.629 $\pm$ 0.124          \\
			GMDLDR          & \textbf{1.000 $\pm$ 0.000} & 0.873 $\pm$ 0.017          & 0.778 $\pm$ 0.036          & 0.955 $\pm$ 0.005          & 0.766 $\pm$ 0.004          \\
			Ligdream        & \textbf{1.000 $\pm$ 0.000} & 0.995 $\pm$ 0.004          & \textbf{0.934 $\pm$ 0.005} & \textbf{1.000 $\pm$ 0.000} & 0.847 $\pm$ 0.002          \\
			REINVENT        & 0.985 $\pm$ 0.002          & \textbf{1.000 $\pm$ 0.000} & 0.763 $\pm$ 0.019          & \textbf{1.000 $\pm$ 0.000} & 0.835 $\pm$ 0.002          \\
			REINVENT 3.0    & 0.989 $\pm$ 0.004          & 0.740 $\pm$ 0.020          & 0.826 $\pm$ 0.027          & 0.886 $\pm$ 0.009          & 0.781 $\pm$ 0.006          \\
			TransVAE        & 0.599 $\pm$ 0.003          & \textbf{1.000 $\pm$ 0.000} & 0.925 $\pm$ 0.003          & \textbf{1.000 $\pm$ 0.000} & \textbf{0.865 $\pm$ 0.000} \\
			Transmol        & \textbf{1.000 $\pm$ 0.000} & 0.687 $\pm$ 0.051          & 0.665 $\pm$ 0.032          & 0.999 $\pm$ 0.000          & 0.808 $\pm$ 0.013          \\ \midrule

			                & \textbf{FCD/Test}          & \textbf{SNN/Test}          & \textbf{Frag/Test}         & \textbf{Scaf/Test}         & \textbf{IntDiv2}           \\ \midrule
			CDN             & 28.748 $\pm$ 5.008         & 0.381 $\pm$ 0.118          & 0.048 $\pm$ 0.034          & 0.202 $\pm$ 0.303          & 0.577 $\pm$ 0.110          \\
			GMDLDR          & \textbf{6.911 $\pm$ 0.490} & \textbf{0.577 $\pm$ 0.013} & 0.872 $\pm$ 0.021          & 0.952 $\pm$ 0.014          & 0.746 $\pm$ 0.004          \\
			Ligdream        & 29.804 $\pm$ 0.877         & 0.230 $\pm$ 0.002          & 0.612 $\pm$ 0.037          & 0.000 $\pm$ 0.000          & 0.838 $\pm$ 0.002          \\
			REINVENT        & 38.210 $\pm$ 0.996         & 0.217 $\pm$ 0.003          & 0.498 $\pm$ 0.026          & 0.002 $\pm$ 0.002          & 0.828 $\pm$ 0.002          \\
			REINVENT 3.0    & 7.153 $\pm$ 0.282          & 0.573 $\pm$ 0.012          & \textbf{0.890 $\pm$ 0.014} & \textbf{0.967 $\pm$ 0.007} & 0.756 $\pm$ 0.007          \\
			TransVAE        & 31.555 $\pm$ 0.591         & 0.205 $\pm$ 0.001          & 0.478 $\pm$ 0.018          & 0.000 $\pm$ 0.001          & \textbf{0.860 $\pm$ 0.000} \\
			Transmol        & 11.525 $\pm$ 0.928         & 0.553 $\pm$ 0.027          & 0.827 $\pm$ 0.160          & 0.785 $\pm$ 0.119          & 0.785 $\pm$ 0.011          \\ \midrule
			\multicolumn{6}{c}{{\bfseries mTOR}}                                                                                                                             \\ \midrule
			\textbf{Method} & \textbf{valid}             & \textbf{unique@1000}       & \textbf{Filters}           & \textbf{Novelty}           & \textbf{IntDiv}            \\ \midrule

			CDN             & 0.976 $\pm$ 0.016          & 0.594 $\pm$ 0.092          & 0.872 $\pm$ 0.035          & 0.975 $\pm$ 0.003          & 0.780 $\pm$ 0.024          \\
			GMDLDR          & \textbf{1.000 $\pm$ 0.000} & 0.996 $\pm$ 0.003          & 0.844 $\pm$ 0.134          & 0.999 $\pm$ 0.001          & 0.828 $\pm$ 0.003          \\
			Ligdream        & \textbf{1.000 $\pm$ 0.000} & 0.999 $\pm$ 0.001          & \textbf{0.953 $\pm$ 0.002} & \textbf{1.000 $\pm$ 0.000} & 0.845 $\pm$ 0.000          \\
			REINVENT        & 0.976 $\pm$ 0.003          & \textbf{1.000 $\pm$ 0.000} & 0.819 $\pm$ 0.011          & \textbf{1.000 $\pm$ 0.000} & 0.840 $\pm$ 0.003          \\
			REINVENT 3.0    & 0.986 $\pm$ 0.003          & 0.980 $\pm$ 0.003          & 0.916 $\pm$ 0.008          & 0.885 $\pm$ 0.006          & 0.836 $\pm$ 0.002          \\
			TransVAE        & 0.594 $\pm$ 0.006          & \textbf{1.000 $\pm$ 0.000} & 0.924 $\pm$ 0.006          & \textbf{1.000 $\pm$ 0.000} & \textbf{0.865 $\pm$ 0.001} \\
			Transmol        & \textbf{1.000 $\pm$ 0.000} & 0.749 $\pm$ 0.104          & 0.729 $\pm$ 0.043          & 0.998 $\pm$ 0.000          & 0.840 $\pm$ 0.012          \\ \midrule

			                & \textbf{FCD/Test}          & \textbf{SNN/Test}          & \textbf{Frag/Test}         & \textbf{Scaf/Test}         & \textbf{IntDiv2}           \\ \midrule
			CDN             & 10.617 $\pm$ 1.700         & \textbf{0.664 $\pm$ 0.043} & 0.855 $\pm$ 0.031          & 0.344 $\pm$ 0.125          & 0.737 $\pm$ 0.031          \\
			GMDLDR          & 9.678 $\pm$ 4.301          & 0.362 $\pm$ 0.011          & 0.901 $\pm$ 0.025          & 0.057 $\pm$ 0.043          & 0.818 $\pm$ 0.007          \\
			Ligdream        & 15.756 $\pm$ 0.182         & 0.292 $\pm$ 0.001          & 0.803 $\pm$ 0.003          & 0.001 $\pm$ 0.000          & 0.839 $\pm$ 0.000          \\
			REINVENT        & 25.530 $\pm$ 0.681         & 0.310 $\pm$ 0.003          & 0.832 $\pm$ 0.008          & 0.005 $\pm$ 0.002          & 0.833 $\pm$ 0.003          \\
			REINVENT 3.0    & \textbf{1.819 $\pm$ 0.072} & 0.613 $\pm$ 0.005          & \textbf{0.995 $\pm$ 0.001} & \textbf{0.824 $\pm$ 0.016} & 0.819 $\pm$ 0.002          \\
			TransVAE        & 20.256 $\pm$ 0.215         & 0.270 $\pm$ 0.001          & 0.833 $\pm$ 0.004          & 0.003 $\pm$ 0.001          & \textbf{0.859 $\pm$ 0.001} \\
			Transmol        & 13.894 $\pm$ 1.740         & 0.534 $\pm$ 0.025          & 0.789 $\pm$ 0.056          & 0.209 $\pm$ 0.096          & 0.805 $\pm$ 0.013          \\ \bottomrule
		\end{tabular}
		\label{tab:moses-vdr}
	}
\end{table}

\section{Discussion} \label{sec:Discussion}

In this study, we proposed a new biologically inspired benchmark for evaluating the performance of \emph{de novo} molecular generation methods and for assessing their ability to learn structural features of ligands responsible for the interaction with a specific target.

For our benchmark we carefully designed three distinct target-based reference datasets for the evaluation of generative models: VDR; GABA\textsubscript{A}; and mTOR. Each of the selected targets is well known and widely studied, and thus a significant number of ligands have been identified for all of them. At the same time, there are a number of key differences \emph{between} the reference datasets that we have extracted and curated. Firstly, the number of ligands in the datasets varies substantially: $N=370$ for VDR; $N=256$ for GABA\textsubscript{A}; and $N=4625$ for mTOR (see Table~\ref{tab:moses_metrics} for a quantitative description of reference datasets including MOSES metrics). This difference in ligand number directly affects the training of the generative models: While for some generative models a small number of samples may be sufficient, others may need more data to train a reasonable model. Secondly, the variability of molecular weights differs between ligands of specific targets: While VDR and GABA\textsubscript{A} ligands have large variability, mTOR ligand variability is comparatively small. Thirdly, the IntDiv within ligands of a specific target differs. While VDR and GABA\textsubscript{A} ligands contain similar molecular scaffolds within their respective datasets, mTOR ligands show greater diversity and may thus require a more complex generative model to capture this variability, which in turn required more training data. In summary, this non-exhaustive list of differences between ligand-classes adds additional constraints for a given generative model to be successful at creating a meaningful output across all benchmark datasets. 

Once the generative model has been trained on a subset of ligands for a given target and tasked to create a focused library, this output 
allows us to: i) calculate the number of recreated ligands; 
ii) calculate DTA predictions; and iii) perform molecular docking.

The recreation metric that has been proposed here allows to not only estimate the ability of creating focused libraries for a given generative model but also examines its output from a functional perspective that has direct relevance for the drug discovery process. A generative model that succeeds in recreating a large fraction of known ligands in an out-of-sample fashion is clearly able to model the variability of this compound class accurately. As a result, this makes the generation of other so far unknown binders more likely.

The second metric we proposed is based on the calculation of the average binding affinity of generated molecules. The underlying assumption here is that if a DTA model is able to accurately predict whether an unseen molecule acts as a ligand for a specific target, then it can be employed as a validator of the molecular outputs of generative models. A large range of DTA predictors have been proposed to date. Some of them being target-specific, while others are target-invariant. In this work we have employed a target-specific SVM, a number of related graph-based architectures, such as GraphDTA and in addition proposed a multi-task learning label encoding approach. Predictions of these models were used to rank generative models according to the predicted average binding affinities of their generative outputs. This ranking showed a high level of homogeneity across generative models and can thus be seen as a legitimization of this approach (see also Figure~\ref{fig:many_rankings1}). In addition, the DTA prediction metric led to results, which are in line with those of the recreation metric.

Molecular docking is \emph{de facto} the most accurate computational tool available for predicting the interaction of two molecules and is being used to evaluate ML-based methods for predicting binding affinities~\cite{Maziarka2020}. Since molecular docking provides insight into the quality of drug-target interactions, docking results are a viable approach to further support the reliability of the proposed alternative metrics. Molecular docking was performed for the VDR reference data as well as the generated output sets. As expected, the ligands from the VDR reference set exhibited the lowest average docking score (i.e. strongest affinity binding). However and more interestingly, the same four generative models (CDN, GMDLDR, REINVENT 3.0 and TransMOL) that performed well in the ROOM and DTA prediction metrics also demonstrated low average docking scores for their generated molecules. Please note that the best poses produced by these generative models showed even the lower docking scores than for calcitriol and had very similar conformations. In summary, molecular docking not only served as an alternative and independent validation of the generative models' output, but was also able to concur the results of the two other proposed metrics.

Additionally, we have examined possible relations between ABRAHAM and MOSES metrics. As stated earlier, fragment similarity (Frag/Test), scaffold similarity (Scaf/Test), similarity to a nearest neighbour (SNN/Test) and Fréchet ChemNet distance (FCD/Test) metrics show a similar trend. They thus confirm the recreation metric, which is not surprising considering the fact that these metrics measure various types of similarities between two sets of molecules. For instance, Frag/Test measures the similarity of BRICS fragments~\cite{brics_fragments} present in reference and generated sets. Since training and testing sets used in our study belong to the same ligand set, e.g. VDR, GABA\textsubscript{A}, or mTOR, they contain similar fragments. The more molecules from the testing set a generative model is able to recreate, the higher the similarity between fragments present in the testing set and the generated set will be, as the number of overlapping molecules, hence overlapping fragments, increases.

As mentioned earlier, CDN, GMDLDR, REINVENT 3.0 and Transmol were able to perform well in all three considered metrics. However, LigDream, REINVENT and TransVAE did not perform optimally for the given tasks.
%
One potential reason for the failure of LigDream and REINVENT are the lengths of the generated SMILES strings. Both methods seem to have difficulty in creating molecules with appropriate numbers of heavy atoms (see upper part of Figure~\ref{fig:docking_dist1}). For LigDream, this may be linked to the shortcomings of the long short-term memory (LSTM).
%
For REINVENT the failure in ligand recreation may be explained by the design of the reward function (termed scoring function), which has sparse rewards and thus making the training unstable. Their reward function design takes single molecules as a target instead of using a sequence of molecules. As a result, the scoring function may fail to navigate the agent through chemical space meaningfully.
%
There are multiple potential reasons for the failure of TransVAE. First, training sets might have been too small to estimate entropy values accurately enough and/or to fine-tune the pre-trained models sufficiently. Second, the dataset that the TransVAE model was pre-trained on and our datasets might have conflicting distributions of molecules. In particular, some atoms present in our dataset were absent in the vocabularies of the pre-trained models.
Additionally, TransVAE uses training sets to calculate entropies and select dimensions of the latent space that have high entropy, but this may not be sufficient for extracting meaningful information from the given training sets. For a more in-depth discussion of how models were trained and why some of them have performed sub-optimally for the given metrics, we would like to refer the reader to the supplementary material.

The unique combination of reference datasets and metrics that we proposed for the evaluation of generative models are more closely related to the drug discovery process since ligand-target relations are taken into account. However, our benchmark currently suffers from a number of limitations that need to be addressed: i) While we have carefully designed three diverse reference datasets, there may be other targets, which may enrich the proposed framework. Some of our future work will focus on creating a more comprehensive set of targets, which will add additional constraints for the creation of novel and more powerful generative models; ii) At this time, only the recreation metric as well as multiple ML-based DTA prediction models are available via our GitHub repository\footnote{https://gitlab.com/cheml.io/abraham}
to test any generative model. The reason for molecular docking not being included is that the required software is proprietary and that some expert knowledge is required to run these types of simulations accurately and interpret the results. We are planning to incorporate this feature into our repository in the future using freely available software, such as AutoDock~\cite{AutoDock}; iii) A further limitation of this study lies in the \emph{in silico} nature of the benchmark. While generative models are able to provide meaningful outputs and have shown their potential of actually being able to assist and speedup the drug discovery process~\cite{merk_tuning_2018, merk_novo_2018, bruns_synthetic_2019, zhavoronkov_deep_2019, Grisoni2021,Friedrich2020, Friedrich2021, Chenthamarakshan2022}, to date, only wet-lab experiments can ultimately decide on whether a given target candidate is effective and safe.

During our analysis, we compared the generative outputs in terms of ML-based DTA predictions and molecular docking. While the comparison of these two metrics generally exhibited high concordance, a number of cases were identified where specific molecules showed diverging scores. Examining these molecules in terms of their structural composure revealed that relying on either DTA predictions or molecular docking alone may lead to inaccurate conclusions about potential binding properties of given molecules. We believe this to be a highly important result, in particular, since molecular docking is generally accepted as the gold standard methodology for high-throughput screening during the early stages of the drug discovery process. Instead of trusting a single methodology, we thus propose to rather employ a multi-modal approach, comprised of a range of metrics, such as the ones that have been utilized in this work.

Machine learning-based generative models are currently receiving considerable attention from the scientific community and their usage has led to many interesting insights as well as novel applications in a whole variety of fields, such as computer science, material science, medical science and drug discovery among many others~{\cite{GM_1,GM_2,GM_3}}. However to date, while of paramount importance for their evaluation, benchmarks have not yet received the necessary attention. In this paper we argue that when designing benchmarks for generative models, structural domain knowledge which is functionally task-relevant needs to be incorporated as an evaluation tool. Here, this has been achieved by creating a diverse collection of benchmark datasets, which are composed of ligand sets and their associated targets. By leveraging these reference datasets and their associated metrics, a direct measure of relevance of a generative output can be obtained and thus a new task-associated descriptor of the model itself can be constructed going well beyond a mere successful data generation by the model in a functionally targeted manner. While having successfully adapted this type of framework for applications in drug discovery, the idea of not only examining generative outputs in isolation (or with respect to its input), but rather paying attention to its \emph{functional properties} is applicable to studying and evaluating generative models in general. Such novel benchmarking approaches including structural and domain prior knowledge which focus on functionality will likely lead to generative models which are enabled to create a more directed and thus more practically useful output.

\section*{Conclusions}

In summary, we propose a unique combination of reference datasets and metrics for evaluating generative ML models for molecular generation from a viewpoint of their potential applications in the drug discovery process. As a first indicator, we assessed the ability of models to recreate known binders from three reference molecular datasets associated with their specific biological targets, namely VDR, GABA\textsubscript{A} and mTOR. As a further indicator, we analyzed ML-based predictions of drug-target affinity between the molecular outputs generated by the models which were trained and fine-tuned on our reference datasets and associated biological targets. Importantly, these two indicators were found to correlate well between each other. In other words, generative models that achieved a high recreation rate were also likely to achieve high ranking results for DTA-based predictions. Furthermore, in the case of VDR, the recreation and DTA metrics were generally found to be in agreement with the results of molecular docking. However, we also identified a number of molecules, where DTA predictions diverged from molecular docking scores. By analyzing of these particular cases, we conclude that for high DTA prediction scores, generated molecules need to exhibit a structurally similar backbone and overall size, while for high molecular docking scores the 3D confirmation and the position of critical interacting functional groups is crucial. In summary, we show that both modalities complement each other and thus both need to be considered to make well-informed judgements about the quality of potential drug-like compounds during the discovery phase of the drug design process.

In addition, this study shows that incorporating structural and domain prior knowledge which focus on functionality when designing benchmarks for generative models can not only help with the evaluation of generative models but may also lead to models with more directed and thus more practically useful output.

\section{Methods} \label{sec:Methods}  %

\subsection{Generative models}
To ensure wide applicability and compatibility of our proposed metrics, seven different generative models have been chosen for assessment: CDN; LigDream; GMDLDR; REINVENT; REINVENT 3.0; Transmol and TransVAE. These models use diverse ML architectures including transformers, recurrent neural networks, autoencoders and reinforcement learning. A brief explanation, graphical overview of each generative model's architecture and how they were trained can be found in the supplementary material, while more detailed descriptions can be found in the original articles.

\subsection{Reference datasets for metric evaluation} 

For training and testing generative models we utilized three distinct chemical datasets that contained sets of ligands for VDR, GABA\textsubscript{A} $\alpha$ subunit and mTOR, all of which are well studied and established targets.

	{\bfseries Calcitriol}, the most active secosteroid vitamin D metabolite, exerts its ligand-dependent agonistic gene regulatory effects through VDR which is a member of the nuclear receptor superfamily of Zn-finger transcription factors~\cite{molnar2014}. VDR's ubiquitous tissue expression pattern~\cite{WANG2012123} indicates its crucial role in nearly all physiological processes from calcium/phosphate mineral homeostasis, cellular proliferation and differentiation, as well as in the regulation of immune responses~\cite{norman2006vitamin}. Most of the 4000-5000 synthetic VDR analogs are modifications of the aromatic secosteroid scaffold and/or aliphatic side chain~\cite{Ferdinand2015}. The VDR ligand dataset, which has been manually curated for matching assay type and activity, consists of 370 molecules extracted from various chemical databases including ChEMBL, PubChem and ZINC.

	{\bfseries BZDs} are not naturally occurring ligands, but are a class of synthesized psychoactive drugs. Their chemical scaffold results from fusing a benzene and a diazepine ring~\cite{Mohler2}. Many of the representatives of this class act by binding to an alternative BZD binding site of the GABA\textsubscript{A} receptor, located between the $\alpha1$ and $\gamma2$ subunits and are acting as agonists~\cite{Mohler2}. Thus, BZDs primarily act as anxiolytics, hypnotics, anticonvulsants and muscle relaxants. A targeted substructure search of the BZD scaffold resulted in 256 molecules with structurally highly similar compounds. While the vast majority of the extracted BZDs are binders of the GABA\textsubscript{A} site, only a small fraction lacks binding data.

	{\bfseries mTOR inhibitors}, modulate the function of an evolutionary conserved Serine/Threonine kinase and its protein complexes play a central role in controlling transcription, translation~\cite{CHEN2020112820}, ribosomal synthesis as well as cell growth and size. They can be categorized into four groups: i) derivatives of the originally identified inhibitor - rapamycin (first generation); ii) smaller ATP-competitive inhibitors (second generation); iii) mTOR/PI3K dual inhibitors (second generation); and iv) all newer inhibitors (third generation)~\cite{CHEN2020112820}. mTOR was chosen as a target with a large number of modulators. While these modulators may not directly bind to mTOR, they nevertheless modulate its activity. For simplicity, we will use the term ligand for all reference compounds throughout the manuscript. The dataset which contains 4770 unique human-specific mTOR compounds was extracted from PubChem, ZINC and ChEMBL.\\

{\bfseries Reference dataset representation and its quantitative description}: All mole\-cules were represented in the simplified molecular-input line-entry system (SMILES) format~\cite{weininger1988smiles} and required to pass RDKit validity~\cite{RDKit}. Subsequently, all molecule's isomeric information was removed, they were canonicalized via RDKit and finally all duplicates were removed. For each reference dataset (VDR, GABA\textsubscript{A}, mTOR), 10 random 50:50 splits were created to obtain training and testing sets. To quantitatively characterize all three datasets, averages of MOSES metrics~\cite{moses} were calculated across splits. By comparing the averaged metrics between the train and test splits, the differences and similarities between datasets can be evaluated. Table~\ref{tab:moses_metrics} provides a summary of the MOSES metrics.

\begin{table}[!ht]
	\caption{Quantitative description of reference datasets including MOSES metrics. N. molecules - number of molecules, N. heavy atoms - average number of atoms, N. clusters - number of clusters, FCD/Test - Fréchet ChemNet Distance to the test set, Scaf/Test - scaffold similarity to test set, SNN/Test - nearest neighbour similarity to test set, IntDiv - internal diversity, logP(W1D) - the octanol-water partition coefficient's 1D Wasserstein-1 distance to the test set, MW(W1D) - 1D Wasserstein-1 distance of molecular weights.}
	\resizebox{\textwidth}{!}{
		\centering
		\begin{tabular}{lccc}
			\toprule
			Metric         & VDR                   & GABA\textsubscript{A} & mTOR                  \\ \midrule
			N. molecules   & 370                   & 256                   & 4625                  \\
			N. heavy atoms & 33.9                  & 23.1                  & 33.1                  \\
			N. clusters    & 50                    & 89                    & 949                   \\
			FCD/Test       & 2.9309 $\pm$ 0.2228   & 5.8417 $\pm$ 0.1589   & 1.0353 $\pm$ 0.04     \\
			SNN/Test       & 0.7779 $\pm$ 0.0055   & 0.6693 $\pm$ 0.008    & 0.7506 $\pm$ 0.0026   \\
			Frag/Test      & 0.981 $\pm$ 0.0077    & 0.8831 $\pm$ 0.0109   & 0.9976 $\pm$ 0.0004   \\
			Scaf/Test      & 0.9455 $\pm$ 0.0174   & 0.9687 $\pm$ 0.0097   & 0.8939 $\pm$ 0.0049   \\
			IntDiv         & 0.7517 $\pm$ 0.0055   & 0.743 $\pm$ 0.0027    & 0.8284 $\pm$ 0.0006   \\
			logP(W1D)      & 0.0121 $\pm$ 0.0158   & 0.044 $\pm$ 0.0346    & 0.0038 $\pm$ 0.0038   \\
			MW(W1D)        & 62.6983 $\pm$ 54.4703 & 73.217 $\pm$ 37.1869  & 21.7956 $\pm$ 16.4942 \\ \bottomrule
		\end{tabular}
		\label{tab:moses_metrics}
	}
\end{table}

Internal diversity (IntDiv) measures the dissimilarity \emph{within} a given molecular set using Tanimoto similarity~\cite{tanimoto}. While all three datasets are diverse, mTOR exhibits the highest diversity. Since VDR ligands and BZDs adhere to a limited number of scaffolds, these datasets are expected to have lower IntDiv. While having similar IntDiv, VDR ligands have higher fragment similarity (Frag/Test) and BZDs are more conserved in terms of scaffolds (Scaf/Test). Apart from similarity considerations \emph{within} datasets, an effort was made to construct datasets that also vary \emph{between} each other. This can be assessed by the average number of atoms, molecular weight (MW) distributions and other chemical properties as described by the average 1D Wasserstein-1 distance (W1D) of logP, among others.

\subsection{Generation of molecules}

For generating molecules, a diverse set of seven generative models were employed and the same pipeline was applied for each of them. For each split, the model was trained and/or fine-tuned on the corresponding ligand training subset. This process resulted in 10 sets of generated molecules per model.
Output sets of generated molecules, were reduced in number by filtering them according to their similarity using the combination of Morgan fingerprint representations with radius 2 and length 2048 and Tanimoto similarity. 

Instead of calculating the similarity between each reference compound with respect to the generated ones, the Butina clustering algorithm~\cite{butina1999unsupervised}, with a similarity cutoff parameter of 0.4~\cite{reinvent} was applied to the reference compounds and the obtained cluster centroids were used for similarity calculation. The maximum similarity for each generated compound was then calculated with respect to the centroids. Table~\ref{tab:moses_metrics} shows the number clusters used for the similarity computation.

This process led to same size subsets of generated compounds for each generative method. The following steps summarize the above described procedure:
\begin{enumerate}
	\item[i)] All 10 splits for a given reference dataset and generative model were concatenated;
	\item[ii)] Invalid molecules were removed using RDKit;
	\item[iii)] All generated molecules were canonicalized using RDKit;
	\item[iv)] Duplicates were removed;
	\item[v)] Generated molecules also present in the training set were removed;\\
		{\it This data was subsequently used for the recreation metric (see Sec.~\ref{sec:room})}
	\item[vi)] The molecules were sorted by their average similarity with respect to reference cluster centroids; and
	\item[vii)] Top 400 molecules with the highest similarity were kept for each generative model; \\
		{\it This data was subsequently used for DTA predictions and Molecular Docking (see Sec.~\ref{sec:dta} and~\ref{sec:dock})}
\end{enumerate}

\subsection{Proposed metrics}

\subsubsection{Recreation metric (ROOM)} \label{sec:room}
In this study, we propose a new metric for evaluating the performance of \emph{de novo} molecular generation models. This metric is based on the assumption that the number of recreated ligands is related to the ability of the model to generate molecules with properties similar to those of known binders. Thus high values of this metric suggest the ability of a given generative model to create molecules with a high probability of being ligands for a specific target.

To calculate this metric, two non-overlapping datasets are required: a training and a test dataset, both of which contain known binders. After training the generative model on the training set and generating a set of output molecules, the test set is used to calculate the overlap between the generated and test set molecules. This procedure must be performed several times to ensure consistency, with varying molecular sets allocated to the test and training sets each time. A detailed workflow for examining the recreation rate of known ligands for a given generative model is shown in Algorithm~\ref{alg:recreation}.

\begin{algorithm}
	\caption{ROOM}
	\begin{algorithmic}[1]
		\Require Average recreation rate
		\State Perform 10 splits of known binders into two equal subsets (50:50)
		\State Use the training molecules of the first split for fine tuning the generative model
		\State Bring molecules into canonical form and discard the stereo-isomeric information
		\State Use canonical form of known binders as a seed for the generative model to create a set of generated molecules
		\State Compare the generated molecules to the known binders in the test set and record the overlap
		\State Repeat steps 2 to 5 for ten splits and report average recreation rate
		\State Repeat steps 2 to 6 for each generative model
	\end{algorithmic}
	\label{alg:recreation}
\end{algorithm}

\subsubsection{DTA prediction-based metric} \label{sec:dta}

Another metric we propose in this benchmark focuses on ML-based DTA prediction. Binding affinity is an important characteristic of protein-ligand interaction and a popular metric in virtual screening. Here, DTA predictions are computed between generative molecular outputs and three predefined reference targets (i.e. VDR, GABA\textsubscript{A}, mTOR). In particular, we use a number of ML-based methods for DTA prediction: an SVM~\cite{SVM}, state-of-the-art GraphDTA~\cite{Nguyen684662} and a newly proposed multitask label encoding model (MLT-LE). In addition, we also use three models from the GraphDTA library, namely GAT-GCN, GCNNet and GINConvNet. All models were trained on subsets extracted from BindingDB~\cite{liu2007bindingdb}, which contains 2.3 million binding records.

There are two main approaches for ML-based DTA prediction: i) A classifier is trained for a specific target. Here, all known ligands are used for model estimation and subsequent DTA prediction. For well studied and characterized targets such as VDR, GABA\textsubscript{A}, and mTOR, where a lot of experimental data is available, it is possible to train small but accurate models. ii) Here a target-independent classifier is trained. These types of DTA prediction models are generally more complex, require vast amounts of data for training, but are able to generalize well to new, unseen drug-target pairs with unknown affinity. In this work, we use both approaches to rank the generated molecules.

\paragraph{SVM}
To binarize the extracted binding data, we used a threshold of $<1 \mu M$ to distinguish between active and inactive ligands (as has previously been done in~\cite{Moret2020}). All datasets were class-balanced using the random oversampling technique from the imbalanced-learn package~\cite{imblearn}. For the VDR subset, 4.7\% of the data were excluded, and for GABA\textsubscript{A} and mTOR, 25.7\% and 27.7\%, respectively. After the preprocessing, the VDR, GABA\textsubscript{A}, and mTOR datasets totaled 742, 678, and 7072 records, respectively.
Three SVMs with RBF kernels and automatic per-class parameter balancing were trained in scikit-learn v1.0.1~\cite{pedregosa_scikit-learn_2011}, using the Morgan fingerprint representation with a radius of 2 and a length of 2048. The regularization parameters of these models were tuned using a grid search with 3-fold cross-validation, with $C= 1, 4, 16$ for the VDR, GABA\textsubscript{A}, and mTOR datasets, respectively. The value of $\gamma$ was set to $\frac{1}{N*Var}$, where $N$ is the number of features and $Var$ is the total variation in the feature data. Cross-validation of VDR, GABA\textsubscript{A}, and mTOR classifiers resulted in an accuracy of 88.5\%, 89.4\% and 93.7\%, respectively, and an F1 score of 88.6\%, 88.9\% and 93.6\%. After cross-validation, the classifiers were trained on all data with hyperparameters obtained from cross-validation.

\paragraph{GraphDTA and MLT-LE}
The training datasets for GraphDTA and MLT-LE were obtained from TDC BindingDB~\cite{TherapeuticsDataCommons} and BindingDB v2022m3~\cite{liu2007bindingdb}, respectively. Only human target-specific information from both datasets were retrieved. Both datasets contained multiple binding affinity records for the same drug-target pairs due to different experimental assays. All records for drug-target pairs with a variance of more than two standard deviations were removed from the TDC BindingDB data. Additionally, all invalid SMILES strings from the BindingDB v2022m3 were removed and canonicalized using RDkit. The median was calculated for each drug-target pair with multiple records~\cite{tang_making_2014}. For the TDC BindingDB 42,236, 296,685, 766,904 and 0 unique drug-target pairs were obtained for K\textit{\textsubscript{D}}, K\textit{\textsubscript{I}}, IC\textit{\textsubscript{50}}, and EC\textit{\textsubscript{50}}, respectively. For BindingDB v2022m3 the following unique drug-target pairs were obtained: 39,379, 214,052, 610,502 and 84,487. Three GraphDTA models were then trained separately for K\textit{\textsubscript{D}}, K\textit{\textsubscript{I}}, and IC\textit{\textsubscript{50}} using the TDC BindingDB, while one MLT-LE model was trained on all data at once using BindingDB v2022m3. Further details on the training procedure can be found in the supplementary material. The data were subsequently divided into training and test sets with 80:20 splits.

GraphDTA represents small molecules as a graph and target are encoded by 1D convolutional neural networks (CNNs)~\cite{Nguyen684662}. GraphDTA models were implemented using PyTorch Geometric~\cite{Fey/Lenssen/2019}.
In addition, we incorporate a novel multitask label encoding approach termed MLT-LE, that allows all binding affinity constants, namely K\textit{\textsubscript{D}}, K\textit{\textsubscript{I}}, IC\textit{\textsubscript{50}}, and EC\textit{\textsubscript{50}}, to be used for training simultaneously and in turn predicts them at the same time.

Table~\ref{tab:tab_dta_ref} provides an overview of the dataset, the affinity measures and a summary of all employed architectures for DTA prediction.

\begin{table}[!ht]
	\caption{Standard affinity measures and DTA prediction method reference}
	\renewcommand{\arraystretch}{1.7}
	\resizebox{\textwidth}{!}{
		\begin{tabular}{@{}p{2.3cm}p{5cm}p{5cm}@{}}
			\toprule
			\textbf{Dataset}                                                                                                                                                                                                                                                                                                                                                                                                                                                                                                                                                                        \\ \midrule
			BindingDB                      & \multicolumn{2}{m{10cm}}{BindingDB is a public, web-accessible database of measured binding affinities.}                                                                                                                                                                                                                                                                                                                                                                                                                                               \\ \midrule
			\textbf{Biological Activity Measures}                                                                                                                                                                                                                                                                                                                                                                                                                                                                                                                                                   \\ \midrule

			pK\textit{\textsubscript{D}}   & \multicolumn{2}{m{10cm}}{The equilibrium dissociation constant (K\textit{\textsubscript{D}}) measures the propensity of a ligand-target complex to dissociate. In other words, (K\textit{\textsubscript{D}}) measures the equilibrium between the ligand-target complex and the dissociated components. pK\textit{\textsubscript{D}} is $log$ transformed K\textit{\textsubscript{D}}.}                                                                                                                                                                \\

			pK\textit{\textsubscript{I}}   & \multicolumn{2}{m{10cm}}{The K\textit{\textsubscript{I}} inhibition constant also represents a dissociation constant, but more narrowly for the binding of an inhibitor to an enzyme target. An inhibitor is a ligand whose binding reduces the catalytic activity of the enzyme.  K\textit{\textsubscript{I}} can thus be used as an indicator of how potent the inhibitor is. K\textit{\textsubscript{I}} is $log$ transformed K\textit{\textsubscript{I}}.}                                                                                         \\

			pIC\textit{\textsubscript{50}} & \multicolumn{2}{m{10cm}}{pIC\textit{\textsubscript{50}} stands for a 50 \% inhibitory concentration. It is the concentration of inhibitor required to reduce the biological activity of interest (typically the rate of the enzymatic reaction) to half of the maximum uninhibited value. pIC\textit{\textsubscript{50}} is $log$ transformed pIC\textit{\textsubscript{50}}}                                                                                                                                                                          \\

			pEC\textit{\textsubscript{50}} & \multicolumn{2}{m{10cm}}{EC\textit{\textsubscript{50}} refers to 50 \% effective concentration. It is the ligand concentration at which 50 \% of its maximum effect is exerted on the biological target. The term is rather general and can be applied regardless of whether the ligand inhibits or enhances (induces) the biological function of the target. pEC\textit{\textsubscript{50}} is $log$ transformation of EC\textit{\textsubscript{50}}.}                                                                                                \\

			Proba.                         & \multicolumn{2}{m{10cm}}{Probability estimate of biological activity based on labeling where all measurements $<1 (\mu M)$ were classified as \emph{biologically active}.}                                                                                                                                                                                                                                                                                                                                                                             \\
			\toprule
			\textbf{Architectures}                                                                                                                                                                                                                                                                                                                                                                                                                                                                                                                                                                  \\ \midrule
			GATNet                         & \multirow{3}{3cm}{Models form state-of-the-art DTA library - GraphDTA }                                                                                                                                                                                                                                                                                                                                                                                        & Graph Attention Network~\cite{Nguyen684662}.                                          \\

			GAT-GCN                        &                                                                                                                                                                                                                                                                                                                                                                                                                                                                & Combined Graph Attention Network and Graph Convolutional network~\cite{Nguyen684662}. \\

			GINConvNet                     &                                                                                                                                                                                                                                                                                                                                                                                                                                                                & Graph Iisomorphism Network~\cite{Nguyen684662}.                                       \\

			MLT-LE                         & New deep-learning model                                                                                                                                                                                                                                                                                                                                                                                                                                        & Multi-task Network with Label Encoding.                                               \\

			SVM                            & Classical Support Vector Machine Classifier from scikit-learn package                                                                                                                                                                                                                                                                                                                                                                                          & Classical machine learning classification algorithm.                                  \\
			\bottomrule
		\end{tabular}
		\label{tab:tab_dta_ref}
	}
\end{table}

\subsubsection{Molecular docking} \label{sec:dock}
Molecular docking, a well established tool in virtual screening, was used as a validation methodology for outputs of the generative models as well as for confirming results of ROOM and DTA predictions. The entire VDR reference dataset was used for docking.
Prior to docking, the outputs of each generative model were ranked according to their calcitriol Tanimoto similarity and the top 400 molecules were chosen for docking. After docking, these molecules were ranked using the docking scoring function.
To check the suitability of the generated compounds for molecular docking, the number of heavy atoms was calculated. Based on this analysis only a few outliers were identified and thus the preparation of ligands using the LigPrep module and the subsequent Glide SP molecular docking was performed without further filtering. All procedures were done in Schrödinger Release 2021-2: Maestro, Schrödinger, LLC, New York, NY, 2021.
Prior to docking the molecules from the generative models, the validation of the Glide SP docking procedure was performed by analyzing the top-5 poses produced from the docking of the reference VDR dataset. The analyzed subset contained four nonsecosteridal aryl acetic acid substitution analogs of calcitriol: two isomers of CID11577808 ligand (PDBID 3W0C) and two isomers of CID57432986 ligand (PDBID 3W0A). The last, fifth pose was an isomer of C2$\alpha$-butyl-calcitriol (CID45484898).
A more detailed description of the docking procedure can be found in the supplementary material.

\section*{Acknowledgments}
The co-authors would like to acknowledge the support of Nazarbayev University Research Grants funding to SF, VP and FM (240919FD3926, 11022021FD2903, 110119FD4520). We are also immensely grateful to Prof. Klaus-Robert M\"uller for his valuable comments on a earlier versions of the manuscript. They have increased the quality and scope of the paper substantially. We would also like to thank Alexa Etheridge for proofreading the manuscript.

\bibliography{ms}

\pagebreak

\end{document}


\title[Supplementary Material for ABRAHAM]{Supplementary Material for ``A biologically-inspired evaluation for molecular generative machine learning''}

\author[1,2]{\fnm{Elizaveta} \sur{Vinogradova}}
\equalcont{These authors contributed equally to this work.}

\author[3]{\fnm{Abay} \sur{Artykbayev}}
\equalcont{These authors contributed equally to this work.}

\author[3]{\fnm{Alisher} \sur{Amanatay}}
\equalcont{These authors contributed equally to this work.}

\author[3]{\fnm{Mukhamejan} \sur{Karatayev}}
\author[3]{\fnm{Maxim} \sur{Mametkulov}}
\author[3]{\fnm{Albina} \sur{Li}}
\author[3]{\fnm{Anuar} \sur{Suleimenov}}
\author[3]{\fnm{Abylay} \sur{Salimzhanov}}
\author[1,2]{\fnm{Karina} \sur{Pats}}
\author[3]{\fnm{Rustam} \sur{Zhumagambetov}}
\author*[1]{\fnm{Ferdinand} \sur{Moln\'ar},\email{ferdinand.molnar@nu.edu.kz}}
\author*[4]{\fnm{Vsevolod} \sur{Peshkov},\email{vsevolod.peshkov@nu.edu.kz}}
\author*[3]{\fnm{Siamac} \sur{Fazli},\email{siamac.fazli@nu.edu.kz}}

\affil*[1]{\orgdiv{Department of Biology}, \orgname{Nazarbayev University}, \city{Nur-Sultan}, \country{Kazakhstan}}
\affil*[2]{\orgdiv{Computer Technologies Laboratory}, \orgname{ITMO University}, \city{St. Petersburg}, \country{Russia}}
\affil*[3]{\orgdiv{Department of Computer Science}, \orgname{Nazarbayev University}, \city{Nur-Sultan}, \country{Kazakhstan}}
\affil*[4]{\orgdiv{Department of Chemistry}, \orgname{Nazarbayev University}, \city{Nur-Sultan}, \country{Kazakhstan}}

\date{}
\maketitle

\section{Generative models}
\begin{figure}[ht]
	\centering
	\includegraphics[trim={0 0 0 0},width=11.33 cm]{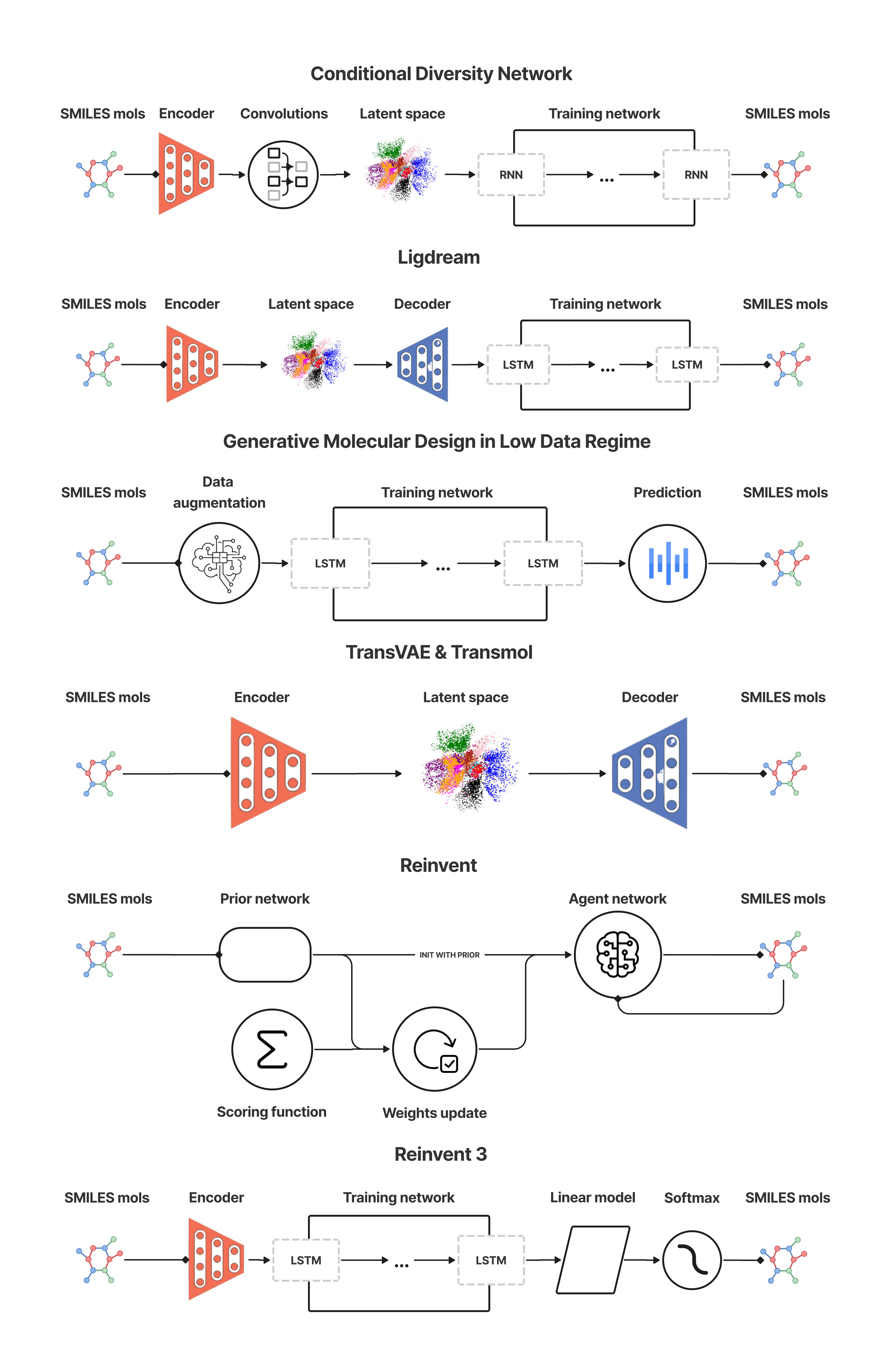}
	\caption{Overview of generative model architectures}
	\label{fig:models}
\end{figure}

A schematic overview of all used generative models can be seen in Figure~\ref{fig:models}.

\subsection{Conditional Diversity Network}
The Conditional Diversity Network (CDN)~\cite{cdn_10.1145/3219819.3219882} is an unsupervised generative algorithm that employs deep learning to generate potential drug-like molecules from seed molecules. It uses a variational autoencoder with an additional ``diversity'' layer to generate sets of molecules similar to the seed. To generate new molecules the model was trained on the train dataset and then the train set molecules were used as seeds for creating the generative output.

\subsection{LigDream}
LigDream~\cite{skalic2019shape} is an algorithm that uses deep learning for generative modelling of molecules. It uses autoencoders to extract shape representations from training set molecules and then uses a mixture of convolutional and recurrent layers to generate SMILES strings from this representation. To generate new molecules the pretrained model, provided by authors, was employed. Training sets were used as seeds to generate new molecules.

\subsection{GMDLDR}
Generative Molecular Design in Low Data Regime (GMDLDR) is a generative model developed by Moret et al.~\cite{Moret2020}. It is based on a recurrent neural network consisting of several layers of long-term short-term memory and batch normalization. It was specifically designed for molecular generation in low-data regimes, i.e., with small sets of molecular input data. This model uses techniques such as data augmentation and transfer learning to adapt to small data regimes. New molecules were generated by predicting subsequent based on the previous ones. Fine-tuning was performed by additional training of the author's pretrained model with custom train/test splits.

\subsection{REINVENT}
REINVENT~\cite{reinvent} enables the navigation of chemical space based on deep Reinforcement Learning (RL). It includes a so-called Prior recurrent neural network (RNN) that is pretrained on molecules from ChEMBL~\cite{chembl}, and an RL agent network with the same architecture as the Prior network. The aim of REINVENT is to effectively use the RL approach to fine-tune the Prior network in order to generate desired molecules based on the scoring function. The key idea of navigation lies in the space of scoring functions, namely a reward function that the agent aims to maximize.

In the original manuscript authors provide $3$ case studies on each of the scoring functions: generation of sulphur molecules, analogues of Celecoxib, and creation active molecules against DRD2 receptor~\cite{reinvent}.

While attempting to generate analogues for the reference datasets, Tanimoto similarity was used as a scoring function individual molecules were used as seeds. We took the pre-trained Prior network weights that were a part of the repository provided by the authors. The target was to fine-tune the agent network to sample only those molecules that have a high Tanimoto similarity to the seed molecule. Tanimoto similarity was acting like a reward function with a threshold: if a sampled molecule and the seed has a similarity greater than the given threshold, the reward was given. For our experiments, we tried different thresholds, but it turned out to be not sensitive to results. After fine-tuning was performed, SMILES were sampled, canonized and checked for validity.

Despite the efforts, no positive results were observable for the REINVENT methodology. The fine-tuned agent was not able to recreate any molecules that were part of the benchmark datasets, namely VDR, BZD and mTOR. We were fine-tuning using the Prior weights that were given by the authors of REINVENT, as well as hyperparameters for training. The fine-tuning was done with the same reward function as proposed by the authors and using the proposed benchmark datasets. Enlarging of sample size of molecules did not help in recreating the molecules, as well as changing the threshold of the reward function. A potential limitation may be the design of the reward function, as it accepts only single molecules as a seed. Hence, the model may not be able to fully utilize the data. There is also a possibility of a covariance shift as the domain of molecules from our dataset differs from the one that it was trained on.

In general, REINVENT~\cite{reinvent} is a powerful, yet limited method of navigating through chemical space. REINVENT highly depends on the reward function, namely the scoring function. We decided to adhere to the original work of the authors and hence, did not change the scoring functions, but there is a chance that with an appropriate scoring function it could be possible to obtain a decent performance on our benchmarks.

\subsection{REINVENT 3.0}

For REINVENT 3.0~\cite{reinvent3}, transfer learning was followed by sampling. The general prior was fine-tuned on training sets of molecules for several epochs. Recommended parameters were used for fine-tuning. The fine-tuned prior was then used to sample the necessary amount of molecules.
REINVENT 3.0 was among the well-performing methods with a comparatively low number of molecules generated. Increasing the number of molecules could potentially increase the absolute numbers for the recreation metric. However, it is also possible that the most efficient generation occurred within the defined limits and does not extend with an increase in sampling. More experiments which are out of the scope of current work are needed to confirm the validity of this hypothesis.

\subsection{Transmol}
Transmol~\cite{transmol_D1RA03086H} is a deep learning method for generation of novel molecules. It employs a variant of the transformer architecture~\cite{attention} to generate focused molecular libraries. Using self-attention layers, Transmol extracts information about molecules and constructs a latent space, which can be later explored using beam search. To generate new molecules a pretrained model was used that was previously published by the authors.

\subsection{TransVAE}

TransVAE~\cite{TransVAE} is a recently developed molecular generative model which employs variational autoencoders (VAE) based on transformers~\cite{attention}. Given an input SMILES string, it first encodes the string into an intermediate latent space and then decodes it back to the SMILES form. The model is trained by minimizing the reconstruction loss between the original SMILES string and its reconstructed version. The key feature of TransVAE is that both its encoder and decoder parts contain attention layers.

New molecules were generated using the $k$-random high entropy dimensions method presented in the original paper, and the model was fine-tuned by additional training on the training parts of the train/test splits.

Unfortunately, both ways of recreating new molecules using TransVAE did not demonstrate any postive results as we were not able to reconstruct even a single molecule from the testing set. Moreover, fine-tuning of the pretrained models did not help either. Even after additional training on the training set for $1000$ epochs, the best result we could achieve was recreating a single molecule from the training set using the one-seed method. One of the potential reasons for failure might be that training sets were too small (only around $100$ molecules for VDR and BZD datasets) to accurately estimate entropy values or to fine-tune the pretrained models sufficiently. Another reason might be that the molecules in VDR and BZD datasets were too different from the molecules in the datasets used to train the pretrained models provided by the authors. Some atoms present in the molecules from VDR and BZD were not present in the vocabularies of the pretrained models, so we had to remove all molecules containing these atoms from VDR and BZD dataset. This reduced the sizes of the training sets further and may have negatively affected the performance of TransVAE.

Overall, even though TransVAE demonstrated excellent results with other metrics, e.g. those in~\cite{moses}, it did not achieve a good score with our set of metrics. Thus, we can conclude that TransVAE was not designed to recreate molecules for focused libraries. However, it is possible that after redesigning or retraining TransVAE for this specific task, it will show good performance in reconstructing focused libraries and achieve high scores our proposed metrics.

\section{DTA prediction models}

\begin{figure}[ht]
	\centering
	\includegraphics[width=\textwidth]{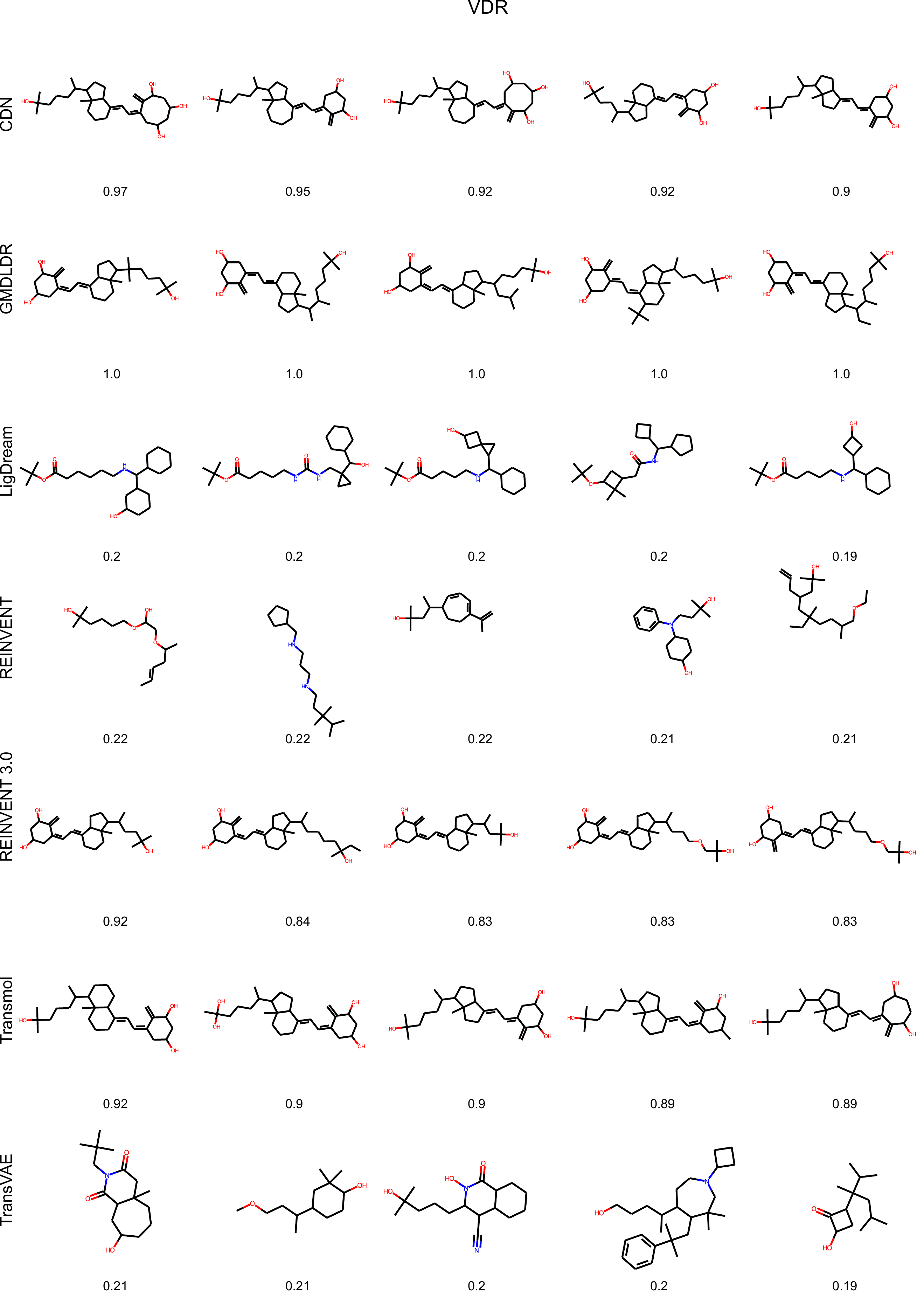}
	\caption{Top 5 structures sorted by Tanimoto Similarity score in the VDR dataset}
	\label{fig:sim_top5_vdr}
\end{figure}

\begin{figure}[ht]
	\centering
	\includegraphics[width=\textwidth]{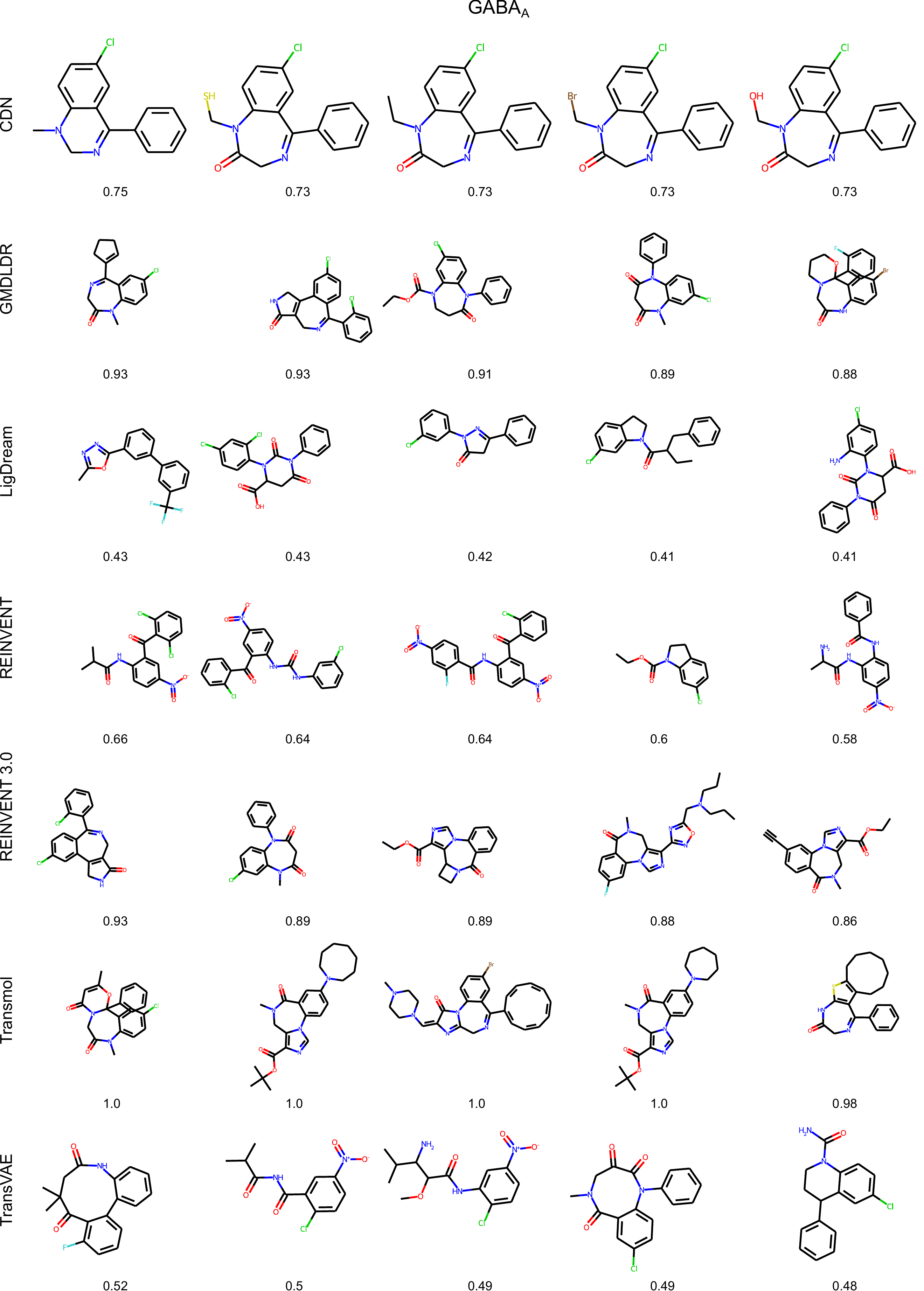}
	\caption{Top 5 structures sorted by Tanimoto Similarity score in the GABA\textsubscript{A} dataset}
	\label{fig:sim_top5_gaba}
\end{figure}

\begin{figure}[ht]
	\centering
	\includegraphics[width=\textwidth]{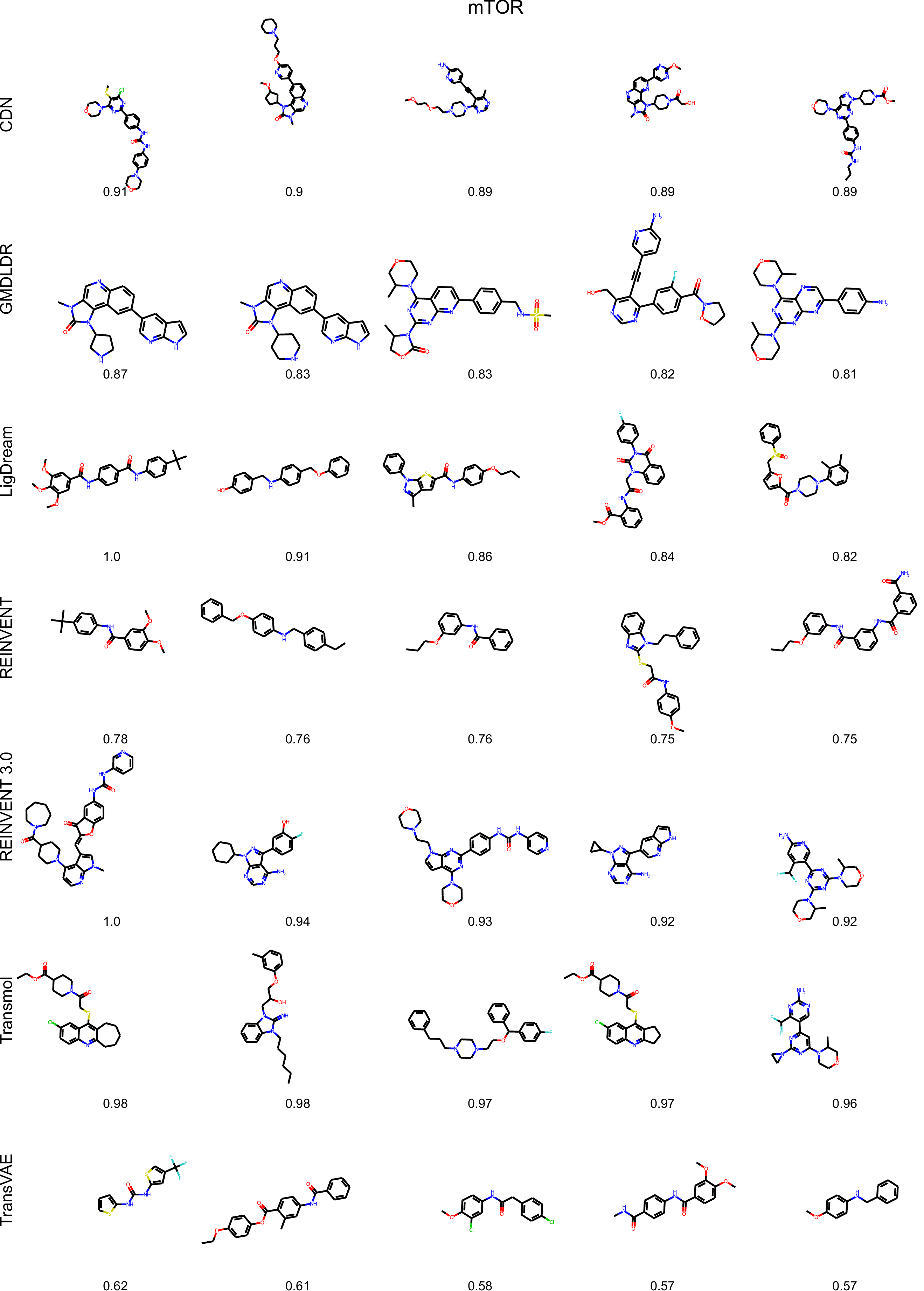}
	\caption{Top 5 structures sorted by Tanimoto Similarity score in the mTOR dataset}
	\label{fig:sim_top5_mtor}
\end{figure}

\begin{figure}[ht]
	\centering
	\includegraphics[width=\textwidth]{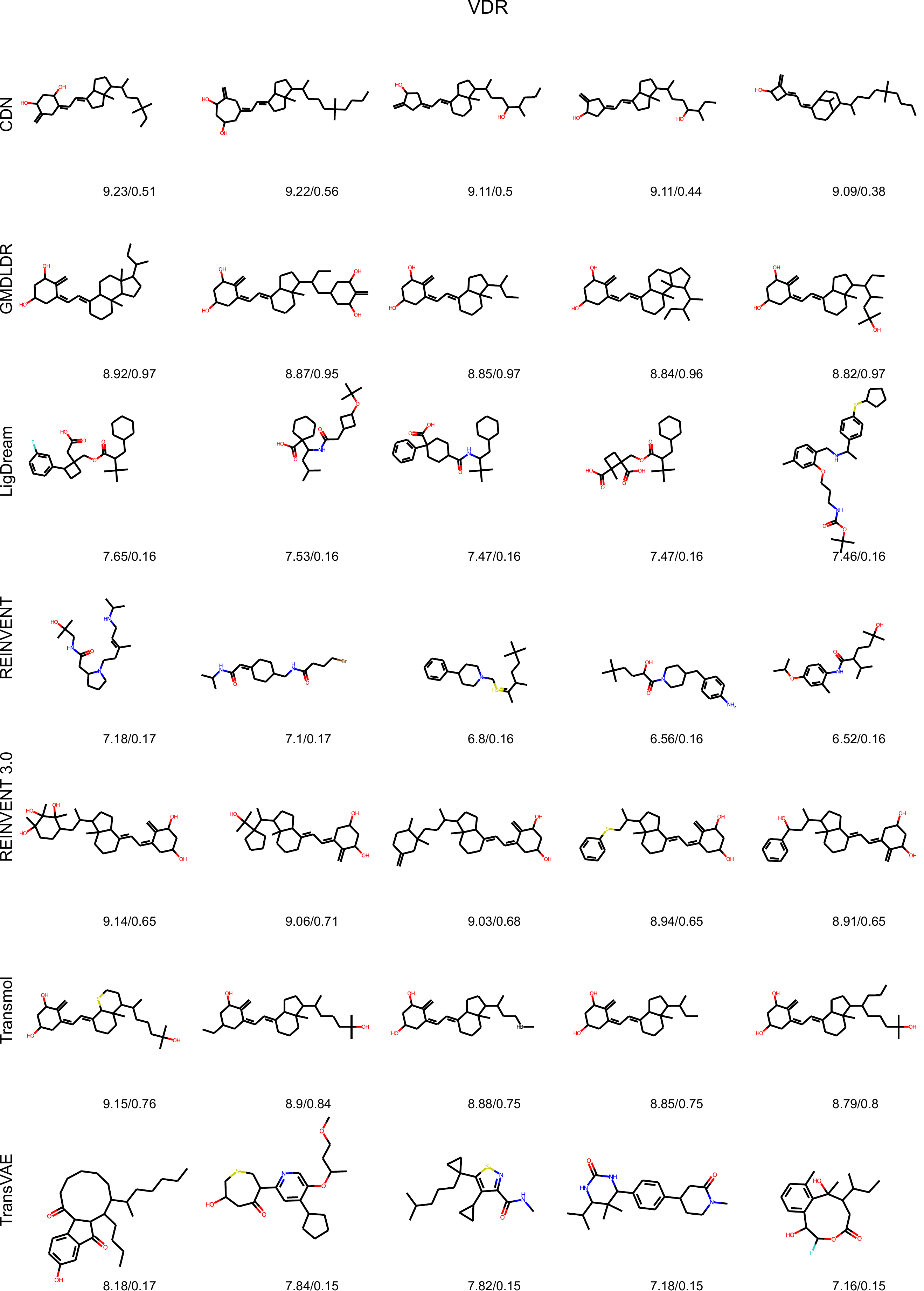}
	\caption{Top 5 structures sorted by GCNNet pK\textit{\textsubscript{D}} score in the VDR dataset}
	\label{fig:gcnnet_top5_vdr}
\end{figure}

\begin{figure}[ht]
	\centering
	\includegraphics[width=\textwidth]{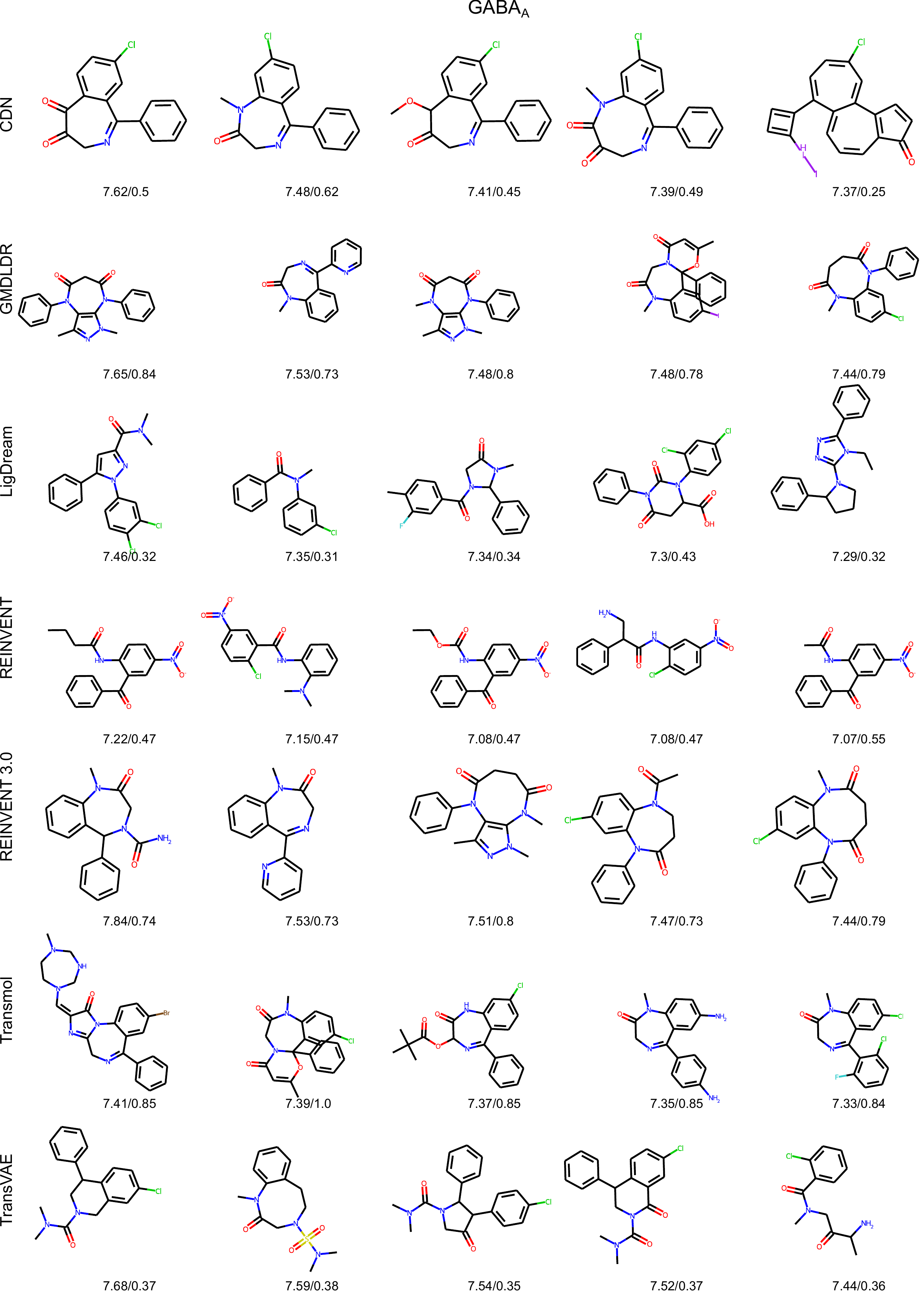}
	\caption{Top 5 structures sorted by GCNNet pK\textit{\textsubscript{D}} score in the GABA\textsubscript{A} dataset}
	\label{fig:gcnnet_top5_gaba}
\end{figure}

\begin{figure}[ht]
	\centering
	\includegraphics[width=\textwidth]{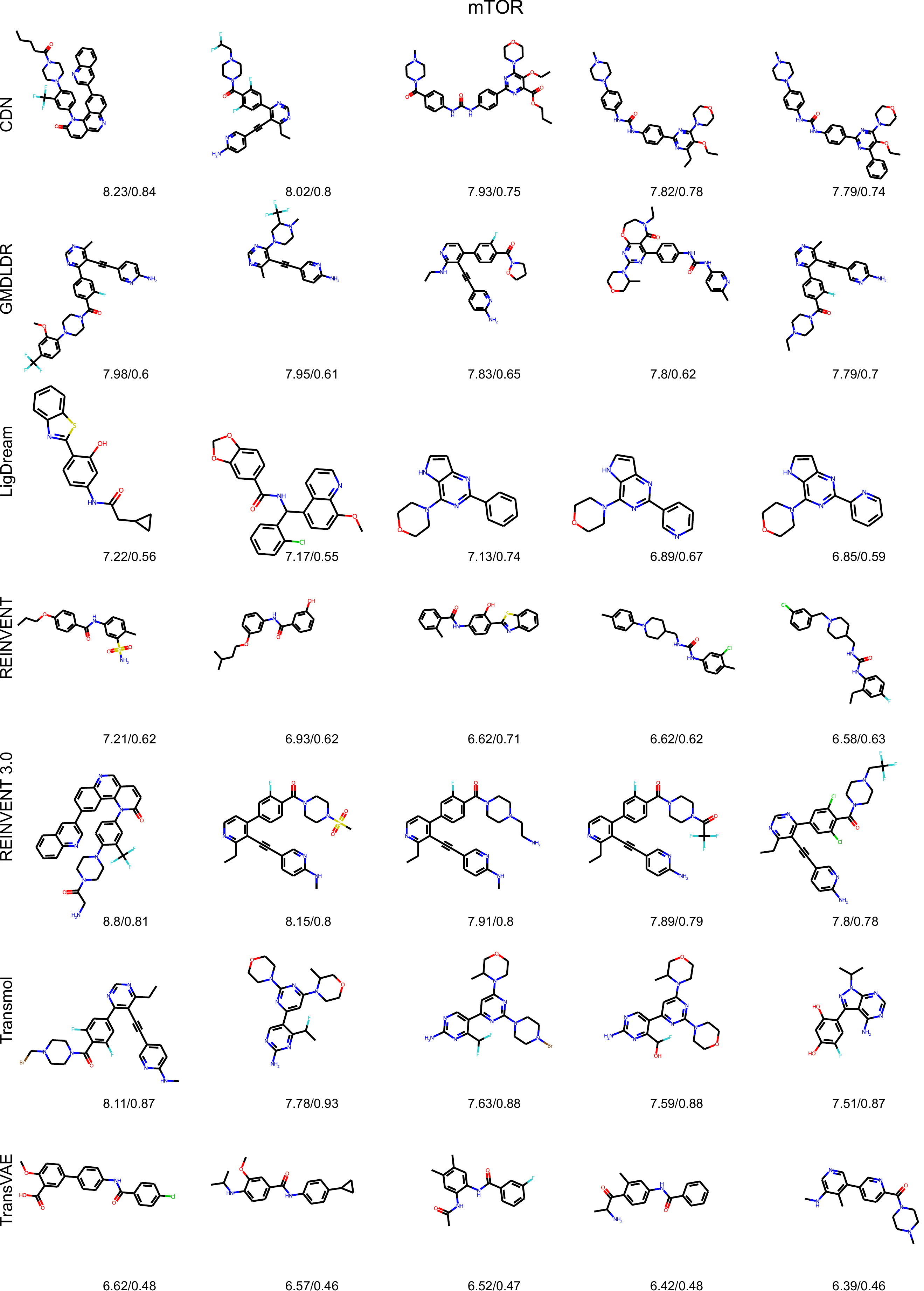}
	\caption{Top 5 structures sorted by GCNNet pK\textit{\textsubscript{D}} score in the mTOR dataset}
	\label{fig:gcnnet_top5_mtor}
\end{figure}

\begin{figure}[ht]
	\centering
	\includegraphics[width=\textwidth]{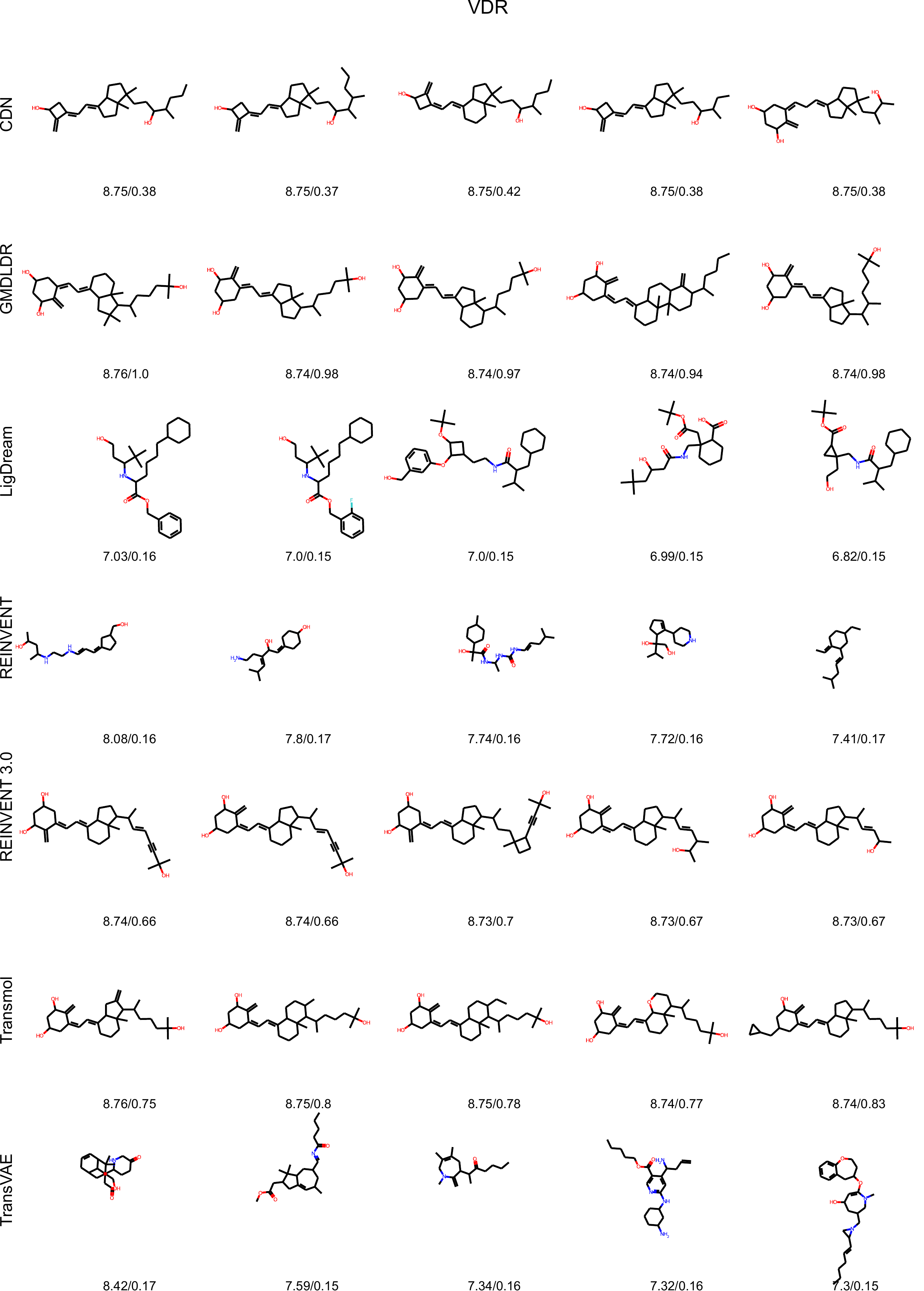}
	\caption{Top 5 structures sorted by MLT-LE pK\textit{\textsubscript{D}} score in the VDR dataset}
	\label{fig:mltle_top5_vdr}
\end{figure}

\begin{figure}[ht]
	\centering
	\includegraphics[width=\textwidth]{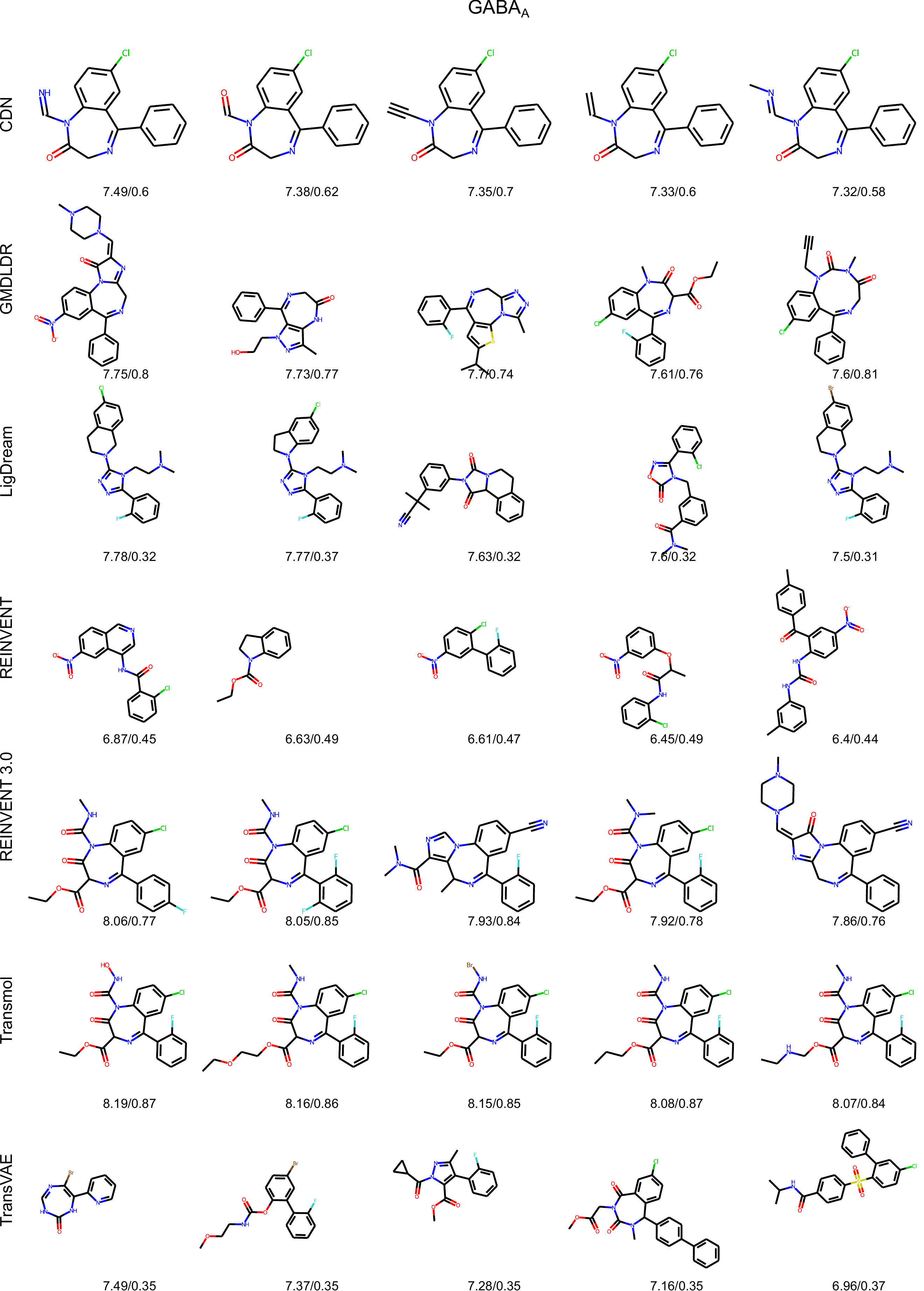}
	\caption{Top 5 structures sorted by MLT-LE pK\textit{\textsubscript{D}} score in the GABA\textsubscript{A} dataset}
	\label{fig:mltle_top5_gaba}
\end{figure}

\begin{figure}[ht]
	\centering
	\includegraphics[width=\textwidth]{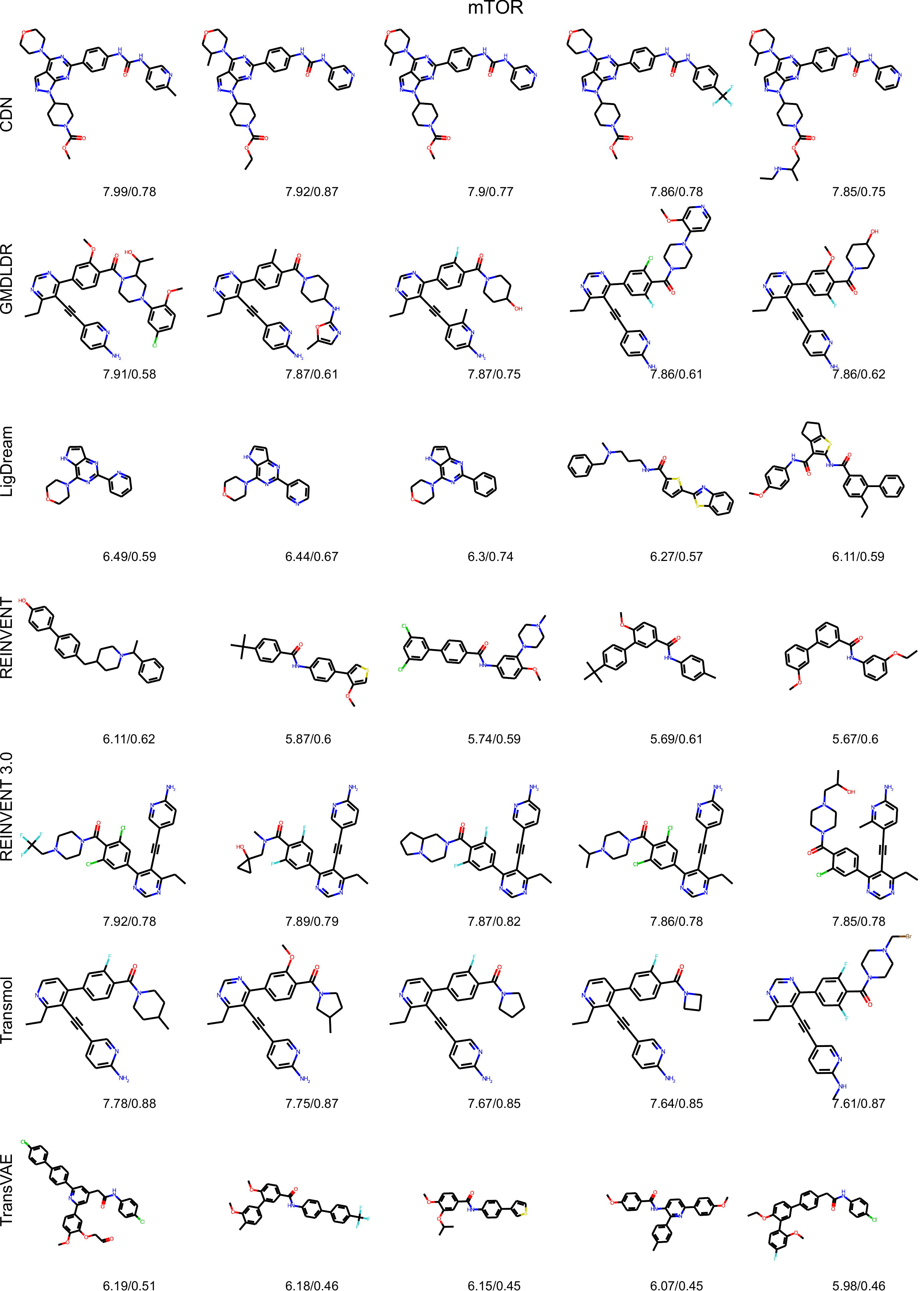}
	\caption{Top 5 structures sorted by MLT-LE pK\textit{\textsubscript{D}} score in the mTOR dataset}
	\label{fig:mltle_top5_mtor}
\end{figure}

The top five generated structures for all generative methods, sorted by their similarity to the training set can be found in Figures~\ref{fig:sim_top5_vdr},~\ref{fig:sim_top5_gaba},~\ref{fig:sim_top5_mtor}, sorted by GraphDTA, GCNNet model pK\textit{\textsubscript{D}} scores in Figures~\ref{fig:gcnnet_top5_vdr},~\ref{fig:gcnnet_top5_gaba},~\ref{fig:gcnnet_top5_mtor} and sorted by MLT-LE pK\textit{\textsubscript{D}} scores in Figures~\ref{fig:mltle_top5_vdr},~\ref{fig:mltle_top5_gaba},~\ref{fig:mltle_top5_mtor}. The two values under each structure in each figure represent the corresponding model score and similarity score, respectively.

\subsection{BindingDB} BindingDB is one of the largest datasets for predicting drug-target affinities (DTA). The latest 2022m3 release includes approx. 2.3 million binding data records for more than 8,000 protein sequences and more than one million drug-like molecules~\cite{liu2007bindingdb}. It also contains proteins from different families, which allows the construction of a robust protein-family-invariant model for DTA prediction. Moreover, compared to other DTA datasets such as Davis and KIBA, BindingDB provides a range of affinity measurements including half-maximum effective concentration (EC\textit{\textsubscript{50}}) and half-maximum inhibitory concentration  (IC\textit{\textsubscript{50}}), inhibition (K\textit{\textsubscript{I}}) and dissociation (K\textit{\textsubscript{D}}) constants as BindingDB combines data from different sources, such as PubChem and ChEMBL and other sources~\cite{liu2007bindingdb}.

\subsection{GraphDTA}
The GraphDTA network uses three consecutive GCN layers with ReLU activation functions. Then, to aggregate the entire graph representation, a global max-pooling layer is implemented. In this work, GraphDTA models were implemented using PyTorch Geometric~\cite{Fey/Lenssen/2019}. First, the data were converted into PyTorch format and the length of the protein sequence ($n$) was set to $1000$. A graph representation of the drug's SMILES and label encoding for the protein sequence was then obtained. Training was performed for $1000$ epochs with a batch size of $512$. An Adam optimization algorithm with a constant learning rate of $0.0005$ was used. Since this is a regression problem, the mean square error (MSE) was used as a loss function. The experiments were performed on an NVIDIA DGX A100 server.

To evaluate the performance of the model, RMSE, MSE, CI, Pearson and Spearman correlation coefficients were computed. The GCNnet method was tested on BindingDB (pK\textit{\textsubscript{D}}) and Davis test sets. As mentioned above, the model was trained on BindingDB (pK\textit{\textsubscript{D}}), so it was tested on the unseen Davis dataset to evaluate its generalizability. We collected unique drug-target pairs from Davis that did not overlap with the training set. Table~\ref{table:graphdta} shows the final results. The data in the table demonstrates that the proposed model exhibits similar MSE and CI results on both datasets.

\begin{table}[!ht]
	\caption{Performance of GraphDTA (GCNnet) model on BindingDB (pK\textit{\textsubscript{D}}) and Davis test sets.}

	\resizebox{\textwidth}{!}{
		\begin{tabular}{@{}ccccccc@{}}
			\toprule
			\textbf{Model}          & \textbf{Dataset}                         & \textbf{RMSE} & \textbf{MSE} & \textbf{CI} & \textbf{Spearman} & \textbf{Pearson} \\ \midrule
			\multirow{2}{*}{GCNnet} & BindingDB (pK\textit{\textsubscript{D}}) & 0.676         & 0.476        & 0.853       & 0.754             & 0.847            \\
			                        & Davis                                    & 0.581         & 0.337        & 0.871       & 0.664             & 0.767            \\ \bottomrule
		\end{tabular}
		\label{table:graphdta}
	}
\end{table}

\textbf{About GraphDTA's pros and cons.} The state-of-the-art GraphDTA network trained on BindingDB outperformed all other methods that were evaluated on the same test sets. The performance of GCNNet and GAT\_GCN models demonstrated the effectiveness of GNNs for learning valuable features from the input. Also, the experiments on generative structures show how well they predict completely unseen data. However, the GraphDTA approach also has some limitations. First of all, the complexity of the network in combination with a large sample size requires a lot of computational resources and training time, which makes the real-world application of such models burdensome. For example, training a single GraphDTA model on the biggest BindingDB\_pI\textit{\textsubscript{50}} dataset took about 10 hours on the NVIDIA DGX A100 server. 
The second drawback of the GraphDTA approach is the underlying architecture. Particularly, the usage of 1-D CNN for protein sequence information. Nowadays, it is a recognized fact that LSTM performs better than 1-D CNN to capture the sequential representation. According to the experiment, replacing the 1-D CNN with Bi-LSTM while leaving the remaining architecture of the GraphDTA unchanged, considerably improves the performance of the model~\cite{mukherjee2022deep}. This indicates that Bi-LSTMs may be much more effective for this tasks than 1-D CNN.

More detailed results for GraphDTA and GCNnet pK\textit{\textsubscript{D}} models can be seen in Figures~\ref{fig:hist_gcnnet_pkd_vdr},~\ref{fig:hist_gcnnet_pkd_gaba},~\ref{fig:hist_gcnnet_pkd_mtor}. These figures show the histograms for the distribution of scores for each generative method, as well as one of the structures with the highest scores.



\begin{figure}[ht]
	\centering
	\includegraphics[width=.5\textwidth]{suppl/long_reports/vdr/CDN_400_sorted_GCNNet_BindingDB_pKd_distr_plot.pdf}\hfill
	\includegraphics[width=.5\textwidth]{suppl/long_reports/vdr/Ligdream_400_sorted_GCNNet_BindingDB_pKd_distr_plot.pdf}
	\\[\smallskipamount]
	\includegraphics[width=.5\textwidth]{suppl/long_reports/vdr/LowDataRegime_400_sorted_GCNNet_BindingDB_pKd_distr_plot.pdf}\hfill
	\includegraphics[width=.5\textwidth]{suppl/long_reports/vdr/REINVENTv2_400_sorted_GCNNet_BindingDB_pKd_distr_plot.pdf}
	\\[\smallskipamount]
	\includegraphics[width=.5\textwidth]{suppl/long_reports/vdr/REINVENTv3_400_sorted_GCNNet_BindingDB_pKd_distr_plot.pdf}\hfill
	\includegraphics[width=.5\textwidth]{suppl/long_reports/vdr/TransMol_400_sorted_GCNNet_BindingDB_pKd_distr_plot.pdf}
	\\[\smallskipamount]
	\includegraphics[width=.8\textwidth]{suppl/long_reports/vdr/TransVAE_400_sorted_GCNNet_BindingDB_pKd_distr_plot.pdf}
	\caption{Detailed VDR results for GraphDTA, GCNnet pK$_{D}$ model}
	\label{fig:hist_gcnnet_pkd_vdr}
\end{figure}

\begin{figure}[ht]
	\centering
	\includegraphics[width=.5\textwidth]{suppl/long_reports/benzo/CDN_400_sorted_GCNNet_BindingDB_pKd_distr_plot.pdf}\hfill
	\includegraphics[width=.5\textwidth]{suppl/long_reports/benzo/Ligdream_400_sorted_GCNNet_BindingDB_pKd_distr_plot.pdf}
	\\[\smallskipamount]
	\includegraphics[width=.5\textwidth]{suppl/long_reports/benzo/LowDataRegime_400_sorted_GCNNet_BindingDB_pKd_distr_plot.pdf}\hfill
	\includegraphics[width=.5\textwidth]{suppl/long_reports/benzo/REINVENTv2_400_sorted_GCNNet_BindingDB_pKd_distr_plot.pdf}
	\\[\smallskipamount]
	\includegraphics[width=.5\textwidth]{suppl/long_reports/benzo/REINVENTv3_400_sorted_GCNNet_BindingDB_pKd_distr_plot.pdf}\hfill
	\includegraphics[width=.5\textwidth]{suppl/long_reports/benzo/TransMol_400_sorted_GCNNet_BindingDB_pKd_distr_plot.pdf}
	\\[\smallskipamount]
	\includegraphics[width=.8\textwidth]{suppl/long_reports/benzo/TransVAE_400_sorted_GCNNet_BindingDB_pKd_distr_plot.pdf}
	\caption{Detailed GABA\textsubscript{A} results for GraphDTA, GCNnet pK\textit{\textsubscript{D}} model}
	\label{fig:hist_gcnnet_pkd_gaba}
\end{figure}

\begin{figure}[ht]
	\centering
	\includegraphics[width=.5\textwidth]{suppl/long_reports/mtor/CDN_400_sorted_GCNNet_BindingDB_pKd_distr_plot.pdf}\hfill
	\includegraphics[width=.5\textwidth]{suppl/long_reports/mtor/Ligdream_400_sorted_GCNNet_BindingDB_pKd_distr_plot.pdf}
	\\[\smallskipamount]
	\includegraphics[width=.5\textwidth]{suppl/long_reports/mtor/LowDataRegime_400_sorted_GCNNet_BindingDB_pKd_distr_plot.pdf}\hfill
	\includegraphics[width=.5\textwidth]{suppl/long_reports/mtor/REINVENTv2_400_sorted_GCNNet_BindingDB_pKd_distr_plot.pdf}
	\\[\smallskipamount]
	\includegraphics[width=.5\textwidth]{suppl/long_reports/mtor/REINVENTv3_400_sorted_GCNNet_BindingDB_pKd_distr_plot.pdf}\hfill
	\includegraphics[width=.5\textwidth]{suppl/long_reports/mtor/TransMol_400_sorted_GCNNet_BindingDB_pKd_distr_plot.pdf}
	\\[\smallskipamount]
	\includegraphics[width=.8\textwidth]{suppl/long_reports/mtor/TransVAE_400_sorted_GCNNet_BindingDB_pKd_distr_plot.pdf}
	\caption{Detailed mTOR results for GraphDTA, GCNnet pK\textit{\textsubscript{D}} model}
	\label{fig:hist_gcnnet_pkd_mtor}
\end{figure}

\subsection{MLT-LE}
The trained model showed comparable performance to GraphDTA on the test set (with CI 81\%) and ranked the generative models in the same order as the SVM method.
This model has two inputs. It accepts two label-encoded strings as its input. Label-encoding is performed using dictionaries, which maps  molecules into smiles format and individual symbols of protein strings into unique values. The protein dictionary consists of 25 unique symbols, and the molecular dictionary contains 63 symbols. Their ranges thus being from 1 to 25 and 1 to 63, respectively. This encoding was adapted from DeepDTA~\cite{ozturk_deepdta_2018}.
Encoded strings are transformed using an embedding layers. Embedding is also learned by model training. Next, three 1D convolution layers are applied and the result is merged using max-pooling. The resulting drug and target representations are then merged and the merged representation is transferred to the four dense layers preceeding the 4 outputs. The dense layers are mixed with the dropout layers. Each output consists of three dense layers.
For the regression task a masked mean squared error is used, and for the classification task binary crossentropy was used. The proposed models are implemented in Python 3.9. Related data, pre-trained models and source code are publicly available at \url{https://github.com/VeaLi/MLT-LE} v0.2.

We hypothesize that the use of multitask learning allows for a more comprehensive evaluation of generated molecules, since multitask models for DTA prediction would be less biased toward specific drug-target pairs and compounds, such as agonists or antagonists, which are not evenly represented across the binding constants. The multitask approach has already shown its reliability in recent studies presenting methods for predicting DTA binding strength~\cite{multi-pli, GanDTI}.

\section{Kernel Density Estimation of similarity and DTA predictions}
As mentioned in~\cite{moses, guacamol}, the distribution of properties is a useful tool for visual evaluation of generated structures to investigate various properties such as the estimated octanol-water partition coefficient (estimated LogP)~\cite{moses}, the Synthetic Accessibility Score (SA)~\cite{moses}, the Quantitative Estimation of Drug-likeness (QED)~\cite{moses}, the molecular similarity~\cite{guacamol} as well as DTA predictions of binding strength.
Figure~5 of the main article shows the kernel density estimates calculated with the SciPy package for the distributions of similarity, SVM, GraphDTA, and MLT-LE scores for all generative methods. Plotting these distributions of binding strength scores along with the corresponding distribution of similarity to the training set for the generative methods enables to assess whether a particular generative method tends to replicate the training set and whether, overall, it tends to produce valid and active compounds. At the same time, by comparing several methodologically different approaches for binding strength estimation, it becomes possible to assess the confidence of these estimates.

\section{Molecular docking}
The VDR structure was obtained from Protein Data Bank (PDB id: 1DB1). The structure was refined using Schrodinger Molecular Modeling Suite v2021-4 tools: all missing hydrogen atoms, loops and side chains were restored with Prime module~\cite{Jacobson2004-cn} along with H-bond optimization and energy minimization. The receptor grid was centered on calcitriol, which is the native ligand of the 1DB1 crystal structure.
All libraries were prepared using LigPrep: ten stereoisomers per each SMILES string were generated and all the output molecules were neutralized.
Molecular docking was performed using Glide ~\cite{Friesner2006-ca} with the SP algorithm (standard precision), which is a bit less accurate than XP mode (extra precision) but has a good performance in terms of running time.

\begin{figure}[ht]
	\centering
	\includegraphics[trim={0.1cm 0 0 0},clip,width=\textwidth]{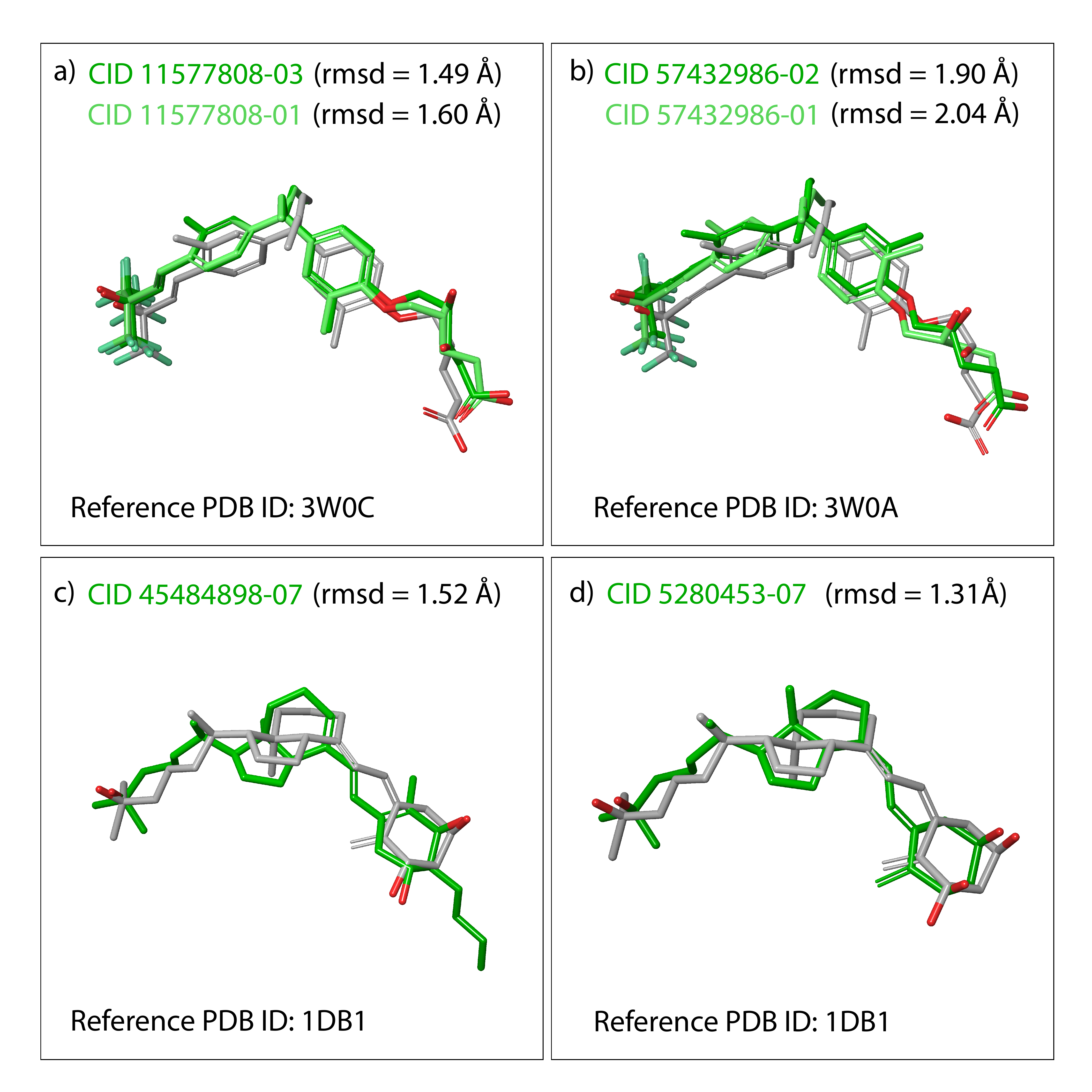}
	\caption{Comparison of the top-5 poses produced by redocking: redocking poses are depicted in green, corresponding PDB reference – in grey.}
	\label{fig:docking_scores_for_top_5}
\end{figure}

Comparison of the top-5 poses produced by redocking are shown in Figure~\ref{fig:docking_scores_for_top_5}. Docking scores for the top-5 seed library poses can be found in Table~\ref{table:docking_table}.

\begin{table}[!ht]
	\caption{Docking scores for top-5 seed library poses: RMSD indicates the difference between the orientation of the seed library ligand and the ligand from the corresponding PDB structure selected as reference}
	\resizebox{\textwidth}{!}{
		\begin{tabular}{@{}ccccccc@{}}
			\toprule
			Pose number & Pubchem CID & Glide SP docking score, kcal/mol & RMSD (in-place), Å & Reference PDB ID \\
			1           & 11577808-03 & -15.97                           & 1.49               & 3W0C             \\
			2           & 11577808-01 & -15.11                           & 1.6                & 3W0C             \\
			3           & 57432986-02 & -15.06                           & 1.9                & 3W0A             \\
			4           & 57432986-01 & -14.68                           & 2.04               & 3W0A             \\
			5           & 45484898-07 & -14.54                           & 1.52               & 1DB1             \\
			344         & 5280453-07  & -12.03                           & 1.31               & 1DB1             \\\bottomrule
		\end{tabular}
		\label{table:docking_table}
	}
\end{table}

\bibliography{supplement}